\documentclass[journal]{IEEEtran}
\usepackage{cite}
\usepackage{multirow}
\ifCLASSINFOpdf
  \usepackage[pdftex]{graphicx}
  \DeclareGraphicsExtensions{.pdf,.jpeg,.png}
\else
  \usepackage[dvips]{graphicx}
  \DeclareGraphicsExtensions{.eps}
\fi

\usepackage{color}

\usepackage{amsmath}
\usepackage{algorithmic,algorithm}
\usepackage{booktabs,multirow}
\makeatletter
\newcommand{\removelatexerror}{\let\@latex@error\@gobble}
\makeatother
\usepackage{pifont}
\usepackage{makecell}

\usepackage{url}
\usepackage[switch, modulo]{lineno}
\usepackage{amsfonts,amssymb,amsthm}

\usepackage{subfigure}
  
\begin{document}
\title{Variational Dynamic for Self-Supervised\\Exploration in Deep Reinforcement Learning}
\author{Chenjia~Bai,
        Peng~Liu,
        Kaiyu~Liu,
        Lingxiao~Wang,
        Yingnan~Zhao,           
        and~Lei~Han
%\thanks{This work was supported in part by the National Natural Science Foundation of China under Grant 51935005, in part by the Fundamental Research Program under Grant JCKY20200603C010, and in part by the Science and Technology on Space Intelligent Laboratory under Grant ZDSYS-2018-02. (\emph{Corresponding author: Chenjia Bai})}
\thanks{Chenjia Bai, Peng Liu, Kaiyu Liu, and Yingnan Zhao are with the Department of Computer Science and Technology, Harbin Institute of Technology, Harbin 150001, China (e-mail: bai\_chenjia@stu.hit.edu.cn; pengliu@hit.edu.cn; hitliukaiyu@foxmail.com; ynzhao\_atari@hit.edu.cn).}
\thanks{Lingxiao Wang is with the Department of Industrial Engineering and Management Sciences, Northwestern University, Evanston, IL 60208, USA (e-mail: lingxiaowang2022@u.northwestern.edu).}
\thanks{Lei Han is with the Tencent Robotics X, Tencent, Shenzhen 518063, China (e-mail: lxhan@tencent.com).}
}
%\thanks{This work has been submitted to the IEEE for possible publication. Copyright may be transferred without notice, after which this version may no longer be accessible.}
\markboth{IEEE Transactions on Neural Networks and Learning Systems}
{Shell \MakeLowercase{\textit{et al.}}: Bare Demo of IEEEtran.cls for IEEE Journals}

\maketitle

\begin{abstract}               % 150-200 word

Efficient exploration remains a challenging problem in reinforcement learning, especially for tasks where extrinsic rewards from environments are sparse or even totally disregarded. Significant advances based on intrinsic motivation show promising results in simple environments but often get stuck in environments with multimodal and stochastic dynamics. In this work, we propose a variational dynamic model based on the conditional variational inference to model the multimodality and stochasticity. We consider the environmental state-action transition as a conditional generative process by generating the next-state prediction under the condition of the current state, action, and latent variable, which provides a better understanding of the dynamics and leads a better performance in exploration. We derive an upper bound of the negative log-likelihood of the environmental transition and use such an upper bound as the intrinsic reward for exploration, which allows the agent to learn skills by self-supervised exploration without observing extrinsic rewards. We evaluate the proposed method on several image-based simulation tasks and a real robotic manipulating task. Our method outperforms several state-of-the-art environment model-based exploration approaches. 

% 158 words
\end{abstract}   

\begin{IEEEkeywords}
variational dynamic model, intrinsic motivation, self-supervised exploration, reinforcement learning.
\end{IEEEkeywords}

\IEEEpeerreviewmaketitle

\section{Introduction}

\IEEEPARstart{R}{einforcement} learning (RL)~\cite{sutton-2018} achieves promising results in solving a wide range of problems, including human-level performance on Atari games~\cite{DQN-2015,ieee-2}, the board game Go~\cite{AlphaGo-2017}, the strategy game StarCraft II~\cite{AlphaStar-2019}, and challenging robotic tasks~\cite{liu2020generating,ieee-3,bai-her}. Most of the successes in RL rely on a well-defined extrinsic reward function from the environment, e.g., a running score from video games. However, in real-world applications, such an extrinsic reward function is sparse or not available, making efficient exploration in such applications a tricky problem.
Conducting exploration without the extrinsic rewards is called the \emph{self-supervised} exploration. From the perspective of human cognition, the learning style of children can inspire us to solve such problems. The children often employ goal-less exploration to learn skills that will be useful in the future. Developmental psychologists consider intrinsic motivation as the primary driver in the early stages of development~\cite{intrinsic-2000}. By extending such idea to RL domain, the `intrinsic' rewards are used in RL to incentivize exploration. Previous formulations of intrinsic rewards used in self-supervised exploration typically utilize `curiosity' corresponding to the prediction-error of environment model~\cite{curiosity-2017,largescale-2019} and the Bayesian uncertainty estimation with ensemble-based~\cite{trials-2018} environment models \cite{disagree-2019}. Both of such formulations require modeling dynamic models of the corresponding environments. 

\begin{figure}[t]
  \centering
  \includegraphics[width=0.25\textwidth]{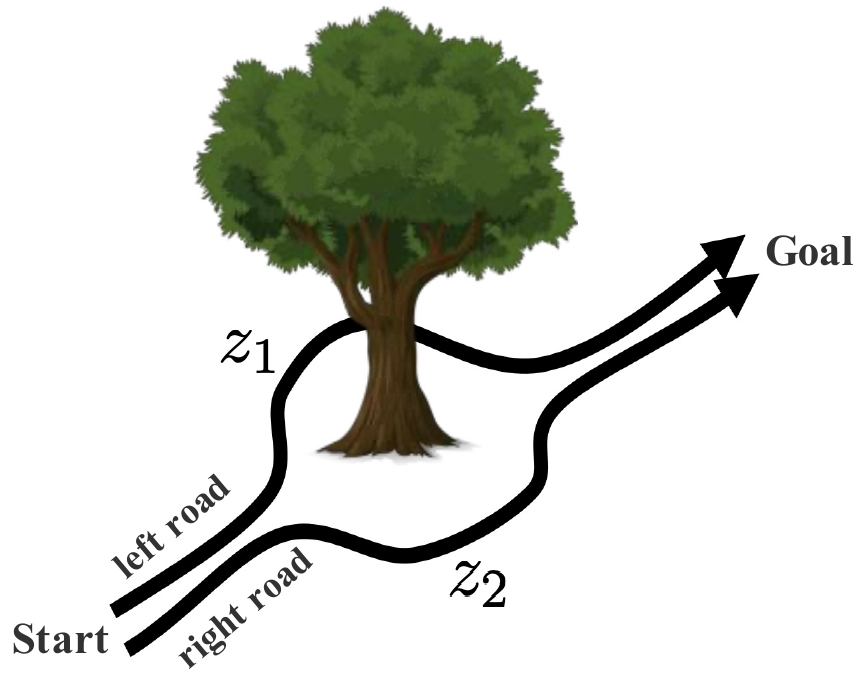}
  \caption{An intuitive example for latent variables in dynamics. We model the multimodality and stochasticity of the dynamics explicitly through latent variables (i.e., $z_1$ and $z_2$) for exploration.}
  \label{fig:tree}
  \vspace{-1.5em}
\end{figure}

An ordinary encoder-decoder based dynamics model that makes deterministic predictions often fails to capture the \emph{multimodality} and \emph{stochasticity} in dynamics and outputs an averaged prediction. An intuitive example is given in Fig.~\ref{fig:tree}, there are two roads (one from the left, and the other from the right) to reach the goal, an ordinary dynamics model will output one pass through the middle. Obviously, the averaged prediction does not reflect the real situation of MDP thus will not generate a reasonable intrinsic reward for the RL agent. However, if we consider the multimodality and stochasticity of the dynamics explicitly through modeling the latent variables (i.e., $z_1$ and $z_2$), then we have a better understanding of the dynamics and lead a better performance in exploration.

\begin{figure}[!t]
\centering
\includegraphics[width=2.5in]{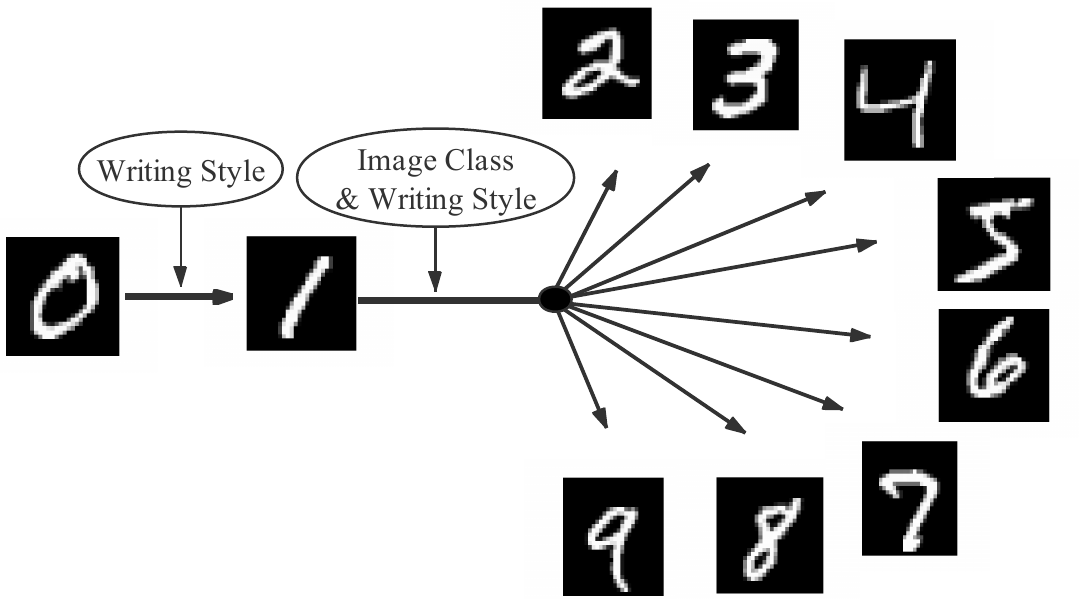}
\caption{MDP of the `Noisy-Mnist'. The state of digit `0' always moves to digit `1', and state of digit `1' moves to other digits with equal probability. The content in the circle represents the latent variables of Noisy-Mnist.}
\label{fig_noise_mnist_mdp}
\vspace{-1.5em}
\end{figure}

For a more complex example, we consider a Markov decision process (MDP) named `Noisy-Mnist'~\cite{disagree-2019} shown in Fig.~\ref{fig_noise_mnist_mdp}, where the state of handwritten digit `0' always moves to handwritten digit `1', and the state of handwritten digit `1' moves to other handwritten digits with equal probability. Although the transition from digit `0' to digit `1' is deterministic, there is stochasticity in the writing styles of a digit, including thickness, slope, and position. For the transition from digit `1' to other digits, modeling the dynamic model needs to consider both the multimodality in different classes of digits and the stochasticity in the writing styles, which is hard for a typical encoder-decoder based neural network. Since the latent variables exist, a typical encoder-decoder network is hard to describe the stochasticity in the transition and outputs blurred predictions. Moreover, although the ensemble-based method~\cite{trials-2018} attempts to measure the stochasticity through training several independent dynamic models, such a method does not consider the multimodality explicitly.

In this paper, we propose the Variational Dynamic Model (VDM), which models the multimodality and stochasticity of the dynamics explicitly based on conditional variational inference. VDM considers the environmental state-action transition as a conditional generative process by generating the next-state prediction under the condition of the current state, action, and latent variable. The latent variable is sampled from a Gaussian distribution to encode the multimodality and stochasticity of the dynamics in a latent space. To conduct efficient exploration based on VDM, we iteratively fit VDM by maximizing the conditional log-likelihood of transitions collected by the agent. To this end, we propose a variational learning objective, which we solve by using stochastic variational inference~\cite{vae-2014,cvae-2015}. The learning of useful latent variables is automatic in VDM training. Through maximizing the learning objective, the latent variables will encode information of multimodality and stochasticity of the underlying dynamics to maximize the log-likelihood of next-state prediction. We do not need to select the latent variables manually, and VDM can be applied in various RL environments and real-world applications.

Upon fitting VDM, we propose an intrinsic reward by an upper bound of the negative log-likelihood, and conduct self-supervised exploration based on the proposed intrinsic reward. We evaluate the proposed method on several challenging image-based tasks, including 1) Atari games, 2) Atari games with sticky actions, which adds more stochasticity in the environment, 3) Super Mario, which we utilize to evaluate the adaptability of VDM to the novel environments, 4) a Multi-player game, which has two controllable agents against each other, and 5) a real robotic manipulating task, which we utilize to evaluate our method in real application scenarios. Experiments demonstrate that VDM outperforms several state-of-the-art dynamics-based self-supervised exploration approaches. 
%
%We use Proximal Policy Optimization (PPO) \cite{ppo-2017} algorithm as the basic policy gradient method. In summary, this paper makes the following contributions.
%\begin{enumerate}
%\item We propose VDM based on conditional variational inference to model the multimodality and stochasticity of the environment explicitly.
%\item We propose an intrinsic reward for self-supervised exploration by an upper bound of the negative log-likelihood of the transition dynamics.
%\item We verify the effectiveness of the proposed intrinsic reward on several challenging image-based tasks.
%\end{enumerate}

\section{Background}

\subsection{Reinforcement Learning}

We consider an MDP that the environment is fully observable, represented as $(\mathcal{S},\mathcal{A},P,r,\rho_0)$. Here $\mathcal{S}$ is the state space, $\mathcal{A}$ is the action space, $P: \mathcal{S}\times \mathcal{A}\rightarrow \mathcal{S}$ is the unknown dynamics model, where $p(s'|s,a)$ is the probability of transitioning to next state $s'$ from current state $s$ by taking action $a$, $r:\mathcal{S}\times \mathcal{A}\rightarrow \mathbb{R}$ is the reward function, and $\rho_0:\mathcal{S}\rightarrow [0,1]$ is the distribution of initial state.

The agent interacts with the environment as follows. In each time step, the agent obtains the current state $s_t$, takes action $a_t$, interacts with the environment, receives the reward $r_t = r(s, a)$, and transits to the next state $s_{t+1}$. The agent selects the action based on the policy $\pi(a_t|s_t)$. The return at time step $t$ is the cumulative discount reward of the future: $R_t=\sum_{i=t}^{T}{\gamma^{i-t} r_i}$, where $\gamma$ is the discount factor. The value function $V^{\pi}(s)$ represents the expected return when starting with the state $s$ and following policy $\pi$ thereafter. The action value function $Q^{\pi}(s,a)$ represents the expected return starting from state $s$, taking action $a$, and following policy $\pi$ thereafter.

\subsection{Policy Gradient}

The goal of RL is to find a policy that maximizes the expected cumulative reward. Policy gradient methods solve RL problem by iteratively following a parameterized policy, sampling data from the parameterized policy, and updating the parameters of policy by policy gradient. The gradient of vanilla policy gradient \cite{sutton-2000} is defined as 
\begin{equation}
\nabla_{\vartheta} J^{{\rm PG}}={\nabla_\vartheta}\mathbb{E}_t[\log\pi_\vartheta(a_t|s_t) A_t],
\end{equation}
where $A_t$ is the advantage function for the current policy defined as $\hat{Q}(s_t,a_t)-b$, $\hat{Q}$ is the estimation of action-value function, and $b$ is the baseline for variance reduction. 

In what follows, we introduce TRPO and PPO algorithms \cite{trpo-2015, ppo-2017}. TRPO updates policy by iteratively maximizing the expected cumulative reward with an extra constraint on KL-divergence between the updated policy and the current policy, which is solved via conjugate gradient algorithm. PPO simplifies the optimization process to first-order methods, and optimizes the policy by stochastic gradient descent. PPO updates policy by the following policy gradient,
\begin{equation}\label{eq-ppo1}
\nabla_{\vartheta} J^{{\rm PPO}}={\nabla_{\vartheta}}\mathbb{E}_t[{{\rm min}(r_t(\vartheta)A_t,{\rm clip}(r_t(\vartheta),1-\epsilon,1+\epsilon)A_t)}],
\end{equation}
where we use $\vartheta$ to represent the parameters of the policy network in PPO, $r_t(\vartheta)=\pi_{\vartheta}(a_t|s_t)/\pi_{\vartheta{{\rm old}}}(a_t|s_t)$, and $\epsilon>0$ is an absolute constant. The clip term ${\rm clip}(r_t(\vartheta),1-\epsilon,1+\epsilon)A_t$ modifies the surrogate objective by removing the incentive for moving $r_t$ outside of $[1-\epsilon,1+\epsilon]$. Then, a minimum of the clipped and unclipped objective is a lower bound (i.e., a pessimistic bound) on the unclipped objective. This clip function is important in PPO. The advantage function $A_t$ in PPO is computed by generalized advantage estimation \cite{gae-2016} as follows,
\begin{equation}
A_t=\delta_t+(\gamma\lambda)\delta_{t+1}+...+(\gamma\lambda)^{T-t+1}\delta_{T},
\end{equation}
where $\gamma$ and $\lambda$ are the hyper-parameters and $\delta_t$ is the TD-error defined as
\begin{equation}
\delta_t=r_t+\gamma V_{\eta}(s_{t+1})-V_{\eta}(s_t).
\end{equation}
Here the value-function network $V_{\eta}$ is updated by following the gradient of the squared TD-error,
\begin{equation}\label{eq-ppo2}
\nabla_{\eta} J^{{\rm PPO}}={\nabla_{\eta}}\mathbb{E}_t[\delta_t]^2,
\end{equation}
where we use $\eta$ to represent the parameters of the value network in PPO. In this paper, we adopt the PPO algorithm with generalized advantage estimation for the proposed self-supervised exploration method.

\subsection{Exploration}

Previous work typically utilizes intrinsic motivation for exploration in complex decision-making problems with sparse rewards. Count-based exploration~\cite{count-2017,count-2016} builds a density model and encourages the agent to visit the states with less pseudo visitation count. Episodic curiosity~\cite{reach-2019} compares the current observation with buffer and uses reachability as the novelty bonus. RND~\cite{RND-2019} measures the state uncertainty by random network distillation. Never give up~\cite{ngu-2020} combines pre-episode and life-long novelty by using an episodic memory-based bonus. Most of these work proposes the final reward for training to characterize the \emph{trade-off} between the extrinsic and intrinsic rewards, which is typically implemented as a linear combination. The intrinsic rewards are crucial when the extrinsic rewards are sparse. 

In this work, we consider self-supervised exploration without extrinsic reward. In such a case, the above trade-off narrows down to a pure exploration problem, aiming at efficiently accumulating information from the environment. Previous self-supervised exploration typically utilizes `curiosity' based on prediction-error of dynamic~\cite{curiosity-2017,aware-2018,largescale-2019} and the Bayesian uncertainty estimation using ensemble-based environment models~\cite{MAX-2019,disagree-2019} or ensemble Q-functions~\cite{bai-1}. Since the agent does pure exploration, the intrinsic motivation becomes the only driving force of the whole learning process. Meanwhile, because the influence of extrinsic rewards is eliminated, the effectiveness of intrinsic rewards can be evaluated independently. After training the pure-exploratory policy with intrinsic rewards, there are several ways to combine the intrinsic policy with extrinsic policies. Scheduled intrinsic drive \cite{sid-2019} uses a high-level scheduler that periodically selects to follow either the extrinsic or the intrinsic policy to gather experiences. MuleX \cite{mulex-2019} learns several policies independently and uses a random heuristic to decide which one to use in each time step. Such policy combination methods perform better than the policy obtained from the linear combination of extrinsic and intrinsic rewards. We focus on developing the pure-exploratory agent and leave the study of policy combination in the future.

The related exploration methods aim to remove the stochasticity of the dynamics rather than modeling it. For example, Inverse Dynamics \cite{curiosity-2017}, Random Features \cite{largescale-2019}, and EMI \cite{emi-2019} learn a feature space to remove the task-irrelevant information in feature space such as white-noise. Curiosity-Bottleneck \cite{kim2019curiosity} and Dynamic Bottleneck \cite{DB-2021} measure reward-relevant novelty through the information bottleneck principle. Contingency awareness \cite{choi2018contingency} builds an attentive model to locate the agent and computes the pseudo-count based on regions around the agent. These methods remove the stochastic part of dynamics to ensure the stability of the intrinsic rewards. In contrast, we propose a novel principle by capturing the multimodality and stochasticity directly through latent space, and measuring the intrinsic reward through sampling latent variables to obtain a tighter upper bound of the true likelihood of dynamics. To the best of our knowledge, a similar problem was only addressed by ensemble-based dynamics in exploration \cite{disagree-2019}. We analyze the ensemble model in Noisy-Mnist and use it as a baseline in experiments.

%The ensemble-based model partly captures the stochasticity of the environment, and VDM performs better through a continuous latent space compared to several individual models in the ensemble. 

%The previous study also establishes methods for exploration beyond intrinsic motivation. Noise network~\cite{noise2-2018} and parameter space noise~\cite{noise1-2018} inject noise directly in the actor-critic network to enhance exploration. Novelty learner~\cite{diversity-2018,diversity-2019} encourages the agent to learn diverse policies that are different from the previously learned policies. Meta-learning intrinsic reward~\cite{onlearning-2018,zheng2019can} learns the optimal reward function by following the gradient of extrinsic rewards. Bootstrapped DQN~\cite{bootstrap-2016,osband-2018} samples $Q$-value from posterior by training several bootstrapped heads of $Q$-network. Based on this method, IDS~\cite{info-2019} performs information-directed exploration with distributional RL~\cite{dis-2017}. OAC~\cite{oac-2019} uses two $Q$-networks to get lower and upper bounds of $Q$-value to perform exploration in continuous control tasks. Hindsight experience replay~\cite{her-2017,entropyher-2019,her-2019} is an exploration method for multi-goal RL.

\subsection{Variational Inference for RL}

Variational inference posits a set of densities and then finds the member in the set that is close to the target \cite{vae-2014,vi-1}. Combining RL and variational inference requires formalizing RL as a probabilistic inference problem \cite{vi-2,vi-3,chen2019variational}. Several RL methods propose to use pseudo-likelihood inference framework \cite{vi-4,vi-5} and expectation maximization (EM) to train policies \cite{vi-6}. VIREL \cite{vi-7} translates the problem of finding an optimal policy into an inference problem. Specifically, VIREL applies EM to induce a family of actor-critic algorithms, where the E-step corresponds to policy improvement and the M-step corresponds to policy evaluation. Dimitri \cite{vi-8} introduces feature-based aggregation to formulate a smaller aggregate MDP where states are related to the features. Xudong, et al. \cite{vi-9} provides probabilistic graphical models to MDP, POMDP, and some important RL methods such as VIME \cite{vime-2016} and variational state tabulation \cite{vi-10}. However, none of these methods focus on modeling multimodality and stochasticity in MDPs. In this paper, we propose to use variational inference to encode the useful information in a latent space and generates intrinsic rewards for efficient exploration in RL.

\section{Variational Dynamic Model for Exploration}

In this section, we introduce VDM for exploration. In section \ref{VDM-s1}, we introduce the theory of VDM based on conditional variational inference. In section \ref{VDM-s2}, we present the detail of the optimizing process. In section \ref{VDM-s3}, we analyze the result of VDM used in `Noisy-Mnist' that models the multimodality and stochasticity in MDP. In section \ref{VDM-s4}, we present the method for calculating a tighter upper bound of the negative log-likelihood of transition as the intrinsic reward. 

\subsection{Variational Dynamic Model}\label{VDM-s1}

Modeling transition dynamics faces two major challenges in many real-world applications: (i) states are high-dimensional, e.g., images, and (ii) the environment contains multimodality and stochasticity, which is hard to model via typical generative models. To this end, we develop VDM, a deep conditional generative model for dynamic modeling in the multimodal and stochastic environment using Gaussian latent variables. VDM considers the state-action transition as a conditional generative process by generating the next-state prediction under the condition of the current state $s$, action $a$, and a latent variable $\mathbf{z}$, which encodes the multimodality and stochasticity in such a generative process. For example, in `Noisy-Mnist', the latent variable $\mathbf{z}$ encodes the class and style of handwritten digits. By changing the latent variables $\mathbf{z}$, the mode and stochasticity of the model will be changed accordingly. In VDM, we generate $\mathbf{z}$ by a diagonal Gaussian distribution. Given $s$, $a$, and a sampled $\mathbf{z}$, VDM generates the next-state prediction $\hat{s}'$ according to $p(s'|s,a,\mathbf{z})$. 

\begin{figure*}[!t]
\centering
\includegraphics[width=6.0in]{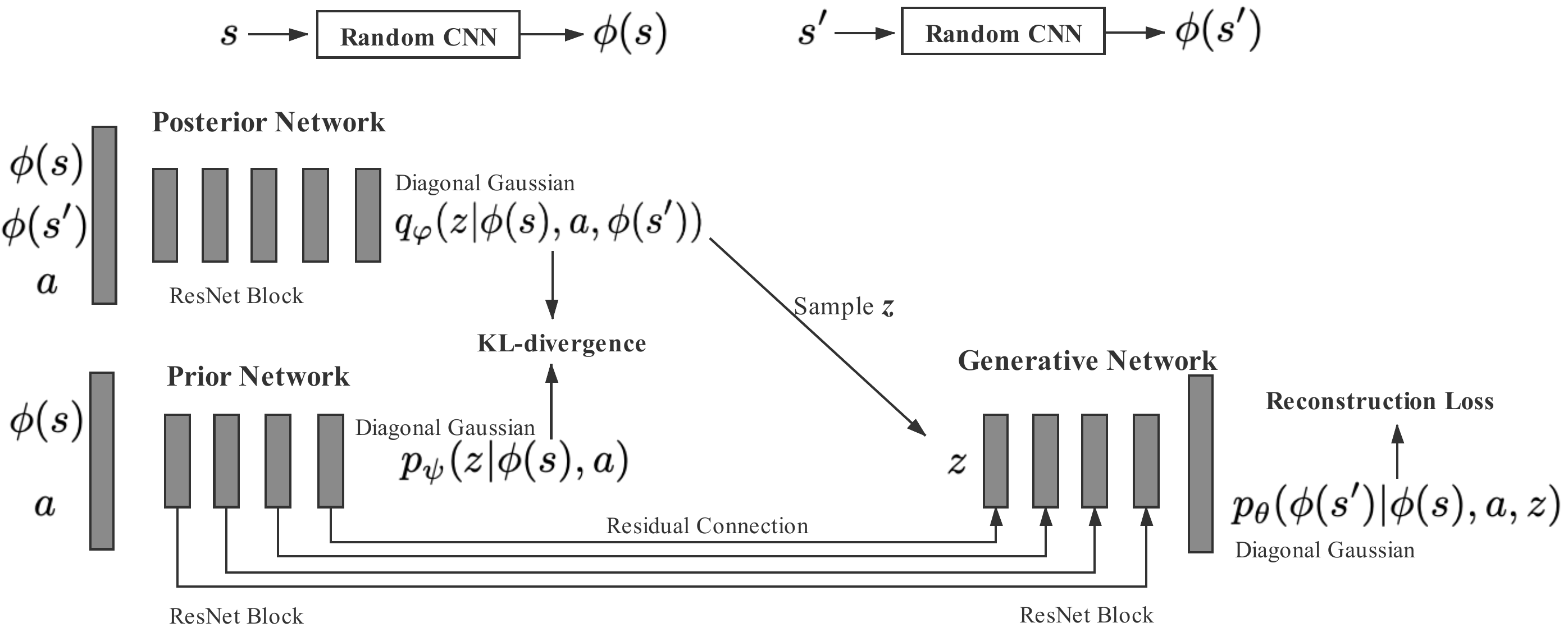}
\caption{VDM architecture. The model contains a posterior network, a prior network and a generative network. The diagonal Gaussian is used as the output of each network. The objective function $L_{\rm VDM}$ contains KL-divergence and reconstruction loss. A random CNN is used for feature extraction.}
\label{fig-model-arc}
\vspace{-1em}
\end{figure*}

To train VDM, we maximize the conditional log-likelihood of the fitted transition dynamics. Since the log-likelihood with the latent variable is intractable, we propose a lower bound of the log-likelihood of transition $\log p(s'|s,a)$ and optimize such a lower bound by stochastic gradient variational Bayes. More specifically, we sample the latent variable $\mathbf{z}$ from a posterior distribution $\mathbf{z}\sim q(\mathbf{z}|s,a,s')$. The log-likelihood objective of the transition is as follows,
\begin{equation}\label{eq-jensen}
\begin{split}
\log p(s'|s,a)= & \log \mathbb{E}_{q(\mathbf{z}|s,a,s')}\Bigl[\frac{p(s',\mathbf{z}|s,a)}{q(\mathbf{z}|s,a,s')}\Bigr]\\
 {{\geq}}&\:\mathbb{E}_{q(\mathbf{z}|s,a,s')}\Bigl[\log \frac{p(s',\mathbf{z}|s,a)}{q(\mathbf{z}|s,a,s')}\Bigr].
\end{split}
\end{equation}
The inequality in \eqref{eq-jensen} follows from Jensen's inequality. We further decompose the numerator $p(s',\mathbf{z}|s,a)$ in \eqref{eq-jensen} as 
\begin{equation}\label{eq-numerator}
p(s',\mathbf{z}|s,a)=p(s'|s,a,\mathbf{z})p(\mathbf{z}|s,a).
\end{equation}
Plugging \eqref{eq-numerator} into \eqref{eq-jensen}, we obtain that
\begin{equation}\label{eq:111}
\log p(s'|s,a)\:{\geq}\:\mathbb{E}_{q(\mathbf{z}|s,a,s')}\Bigl[\log p(s'|s,a,\mathbf{z})-\log \frac{q(\mathbf{z}|s,a,s')}{p(\mathbf{z}|s,a)} \Bigr].
\end{equation}
The first term in the expectation on the right-hand side of \eqref{eq:111} is the log-likelihood in predicting the next state $s'$ with $\mathbf{z}$. The second term is the \emph{KL-divergence} between the \emph{posterior} $q(\mathbf{z}|s,a,s')$ and the \emph{prior} $p(\mathbf{z}|s,a)$ of latent variable $\mathbf{z}$. We denote by $L_{\rm VDM}$ the lower bound of log-likelihood $\log p(s'|s,a)$ on the right-hand side of \eqref{eq:111} and re-write \eqref{eq:111} as 
\begin{equation}
\log p(s'|s,a)\:{\geq}\:L_{{\rm VDM}},
\end{equation}
where
\begin{equation}\label{eq-lb}
\begin{aligned}
L_{\rm VDM}=\mathbb{E}_{q(\mathbf{z}|s,a,s')}&[\log p(s'|s,a,\mathbf{z})]\\
&-D_{\rm KL}[q(\mathbf{z}|s,a,s') \| p(\mathbf{z}|s,a)].
\end{aligned}
\end{equation}
Since the log-likelihood $\log p(s'|s,a)$ in intractable, to fit VDM, we maximize the variational lower bound $L_{\rm VDM}$ defined in \eqref{eq-lb}.

\subsection{Optimizing Process}\label{VDM-s2}

Following from \eqref{eq-lb}, maximizing the objective function $L_{\rm VDM}$ yields maximizing the log-likelihood of next-state prediction $\mathbb{E}_\mathbf{z}[\log p(s'|s,a,\mathbf{z})]$ with latent variable $\mathbf{z}$, and, meanwhile, minimizing the KL-divergence between posterior $q(\mathbf{z}|s,a,s')$ and prior $p(\mathbf{z}|s,a)$. We use three neural networks parameterized by $\varphi,\psi,\theta$ respectively to implement the probability distributions in \eqref{eq-lb} as follows,
\begin{equation}\nonumber
\begin{split}
{\rm Posterior~network}&:~{q_{\varphi}(\mathbf{z}|s,a,s')} \\
{\rm Prior~network}&:~{p_{\psi}(\mathbf{z}|s,a)} \\
{\rm Generative~network}&:~{p_{\theta}(s'|s,a,\mathbf{z})}
\end{split}
\end{equation}
To maximize the objective in \eqref{eq-lb}, we solve the following maximization problem,
\begin{equation}
(\varphi,\psi,\theta)=\arg\max_{\varphi,\psi,\theta}L_{\rm VDM}(s,a,s';\varphi,\psi,\theta).
\end{equation}
We illustrate the optimization process in Fig.~\ref{fig-model-arc}. Recall that the \textbf{first term} in $L_{\rm VDM}$ is $\mathbb{E}_{q_{\varphi}(\mathbf{z}|s,a,s')}[\log p_{\theta}(s'|s,a,\mathbf{z})]$, which corresponds to the log-likelihood of the next-state prediction condition on the current state, action, and a latent variable. This term cannot be computed directly since the expectation involves the parameter ${\varphi}$ of the posterior. To this end, we adopt the reparameterization trick \cite{vae-2014}, which move the sampling process to the input layer of the network. We denote the distribution of the posterior network by $q_{\varphi}(\mathbf{z}|s,a,s')=\mathcal{N}(\mu_0(s,a,s';\varphi),\Sigma_0(s,a,s';\varphi))$, where $\mu_0$ and $\Sigma_0$ are outputs of the posterior network with parameters $\varphi$. We sample $\mathbf{z}$ from $\mathcal{N}(\mu_0,\Sigma_0)$ by first sampling $\mathbf{\epsilon}\sim \mathcal{N}(\mathbf{0},\mathbf{I})$, then computing $\mathbf{z}=\mu_0+{\Sigma}^{1/2}_0*\epsilon$. The first term of $L_{\rm VDM}$ used for training then takes the form of
\begin{equation}\label{eq-reparameter}
\mathbb{E}_{\mathbf{\epsilon}\sim \mathcal{N}(\mathbf{0},\mathbf{I})}\big[\log p_{\theta}(s'|s,a,\mathbf{z}=\mu_0+{\Sigma}^{1/2}_0*\epsilon)\big],
\end{equation}
where the output of $p_{\theta}$ follows a diagonal Gaussian distribution. Note that maximizing the likelihood of $p_{\theta}$ is equivalent with minimizing the reconstruction error of the next state, as shown in Fig.~\ref{fig-model-arc}. 

Meanwhile, recall that the \textbf{second term} in $L_{\rm VDM}$ is the KL-divergence $D_{\rm KL}[q_{\varphi}(\mathbf{z}|s,a,s') \|p_{\psi}(\mathbf{z}|s,a)]$ between the posterior and prior. We denote the prior by $p_{\mathbf{\psi}}(\mathbf{z}|s,a)=\mathcal{N}(\mu_1(s,a;\psi),\Sigma_1(s,a;\psi))$, where $\mu_1$ and $\Sigma_1$ are outputs of the prior network with parameters $\psi$. Under such parameterization, the KL-divergence in $L_{\rm VDM}$ can be computed explicitly as follows,
\begin{equation}
\label{eq-second-kl}
\begin{split}
&D_{\rm KL}[\mathcal{N}(\mu_0,\Sigma_0)\|\mathcal{N}(\mu_1,\Sigma_1)]=
\frac{1}{2}\big({\rm tr}\big(\Sigma_1^{-1}\Sigma_0\big)\\&+({\mu}_1-{\mu}_0)^{\top}{\Sigma}^{-1}_1(\mu_1-\mu_0)-c+\log\big({|\Sigma_1|}/{|\Sigma_0|}\big)\big),
\end{split}
\end{equation}
where $c$ is the dimension of the latent variable and $|\Sigma|$ is the determinant of the covariance matrix $\Sigma$. 

To perform stochastic gradient ascent to maximize $L_{\rm VDM}$, we use the diagonal Gaussian as the output distribution of each neural networks. Parameterizing the latent space by diagonal Gaussian is widely applied in variational inference to solve complex tasks (e.g., Neural Machine Translation~\cite{schulz2018stochastic} and Image Inpainting~\cite{vaeac-2019}). In practice, we verify that diagonal Gaussian is sufficient to model the variation of various tasks. 
%More expressive distributions such as normalizing flows~\cite{kingma2016improved} is also applicable to improve VDM.

To further improve the generative model, we include residual connections between the prior network and the generative network. Such an approach benefits the generative model from two aspects. On the one hand, since the features of $(s,a)$ are used in the prior network $p_{\psi}(\mathbf{z}|s,a)$, the residual connections allow generative network $p_{\theta}(s'|s,a,\mathbf{z})$ to reuse these features to avoid duplicate feature extraction. On the other hand, since the latent variable $\mathbf{z}$ is extremely low dimensional, it is unlikely to contain sufficient information to generate $(s,a)$ for the generative network. Including skip-connection provides additional information on $(s,a)$ to the generative network, which allows the latent variable $\mathbf{z}$ to focus on encoding the stochasticity of the dynamics. In addition, we train VDM with on-policy experiences in RL instead of a fixed dataset. The change in experience (e.g., change room in Montezuma's Revenge, dig a tunnel in Breakout), along with the policy change,  prevents the generative model from overfitting and mode collapse.

To evaluate VDM empirically, we focus on tasks with image-based observation. Previous studies show that learning forward dynamic in the embedding space filters out irrelevant information of images, and is more computationally efficient~\cite{curiosity-2017}. Also, the empirical result of~\cite{largescale-2019} shows that random feature preserves sufficient information about the raw signal for exploration, which is stable and does not require training. Hence, in this work, we use a small random CNN with the parameter $\phi$ as the feature extractor. In the training of VDM, we feed the states $s$ and $s'$ into the CNN to obtain the embedding representations $\phi{(s)}\in \mathbb{R}^d$ and $\phi(s')\in \mathbb{R}^d$. Then, we use the tuple $[\phi(s),a,\phi(s')]$ as the input to train VDM in the embedding space. The parameter $\phi$ of the random CNN are randomly initialized and remains unchanged during the training process.

For a specific task, the latent variables in evaluation are the same as those in the training process. The latent space $Z$ is trained to encode the multimodality and stochasticity of the specific task, but not over all the tasks. That is, for a specific task like Mspacman, $Z$ will encode the information of death, rebirth, and directions of ghosts, and use this information to perform dynamic modeling and efficient exploration; otherwise, for a real-world robotic manipulating task, $Z$ will encode the information of robot system and different objects. The latent space is not shared among different tasks.

\begin{figure}[!t]
\centering
\subfigure[]{\includegraphics[width=1.5in]{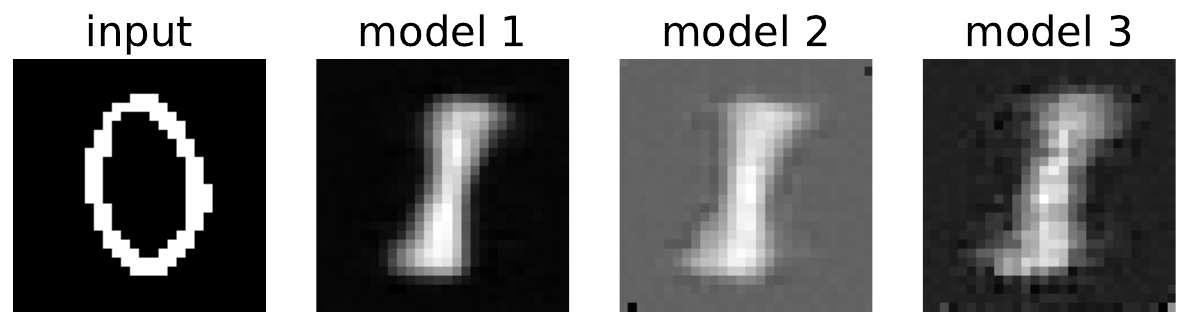}}
\hspace{1em}
\subfigure[]{\includegraphics[width=1.5in]{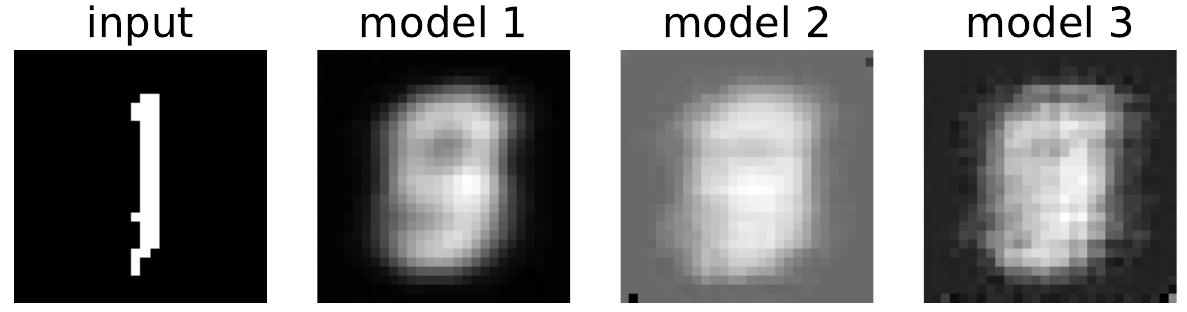}} 
\caption{Result of the probabilistic-ensemble dynamic model in `Noisy-Mnist'. (a) When we input an image of the digit `0', three images are generated from different models. Different models all generate the correct prediction of image class but lacks the diversity of writing styles. (b) When we input an image of the digit `1', the ensemble-based model tends to average the various reasonable predictions and generate blurred images.}
\label{fig_ensemble}
\vspace{-1em}
\end{figure}

\subsection{VDM for Noisy-Mnist}\label{VDM-s3}

In Introduction, the two-roads MDP and Noisy-Mnist illustrate what latent variables can be in specific MDPs. The next-state relies on the latent variables contained in the underlying dynamics. However, we do not need to choose or describe the latent variables manually in practice. Considering the learning objective of VDM in \eqref{eq-lb}, the first-term corresponds to the log-likelihood of the next-state prediction condition on the current state, action, and a latent variable.

(\romannumeral1) Before training, the latent space $Z$ does not contain any useful information since the initial $Z$ is a diagonal Gaussian with random mean and variance. (\romannumeral2) Then, when we maximize the log-likelihood of $\log p(s'|s,a,\mathbf{z})$ by sampling experiences, the prediction almost only considers the state-action $(s,a)$ pair since $\mathbf{z}$ is meaningless. However, because the dynamics contains multimodality and stochasticity, the likelihood the next-state may be hard to maximize since solely relying on $(s,a)$ is insufficient to predict the various possible next-states. (\romannumeral3) In order to further maximize $\mathbb{E}_{\mathbf{z}}[\log p(s'|s,a,\mathbf{z})]$, the model will try to extract useful information (e.g., image class, writing style in Noisy-Mnist) for prediction automatically, and encode these information in the latent space. Since $\mathbf{z}\sim q_{\varphi}(\cdot|s,a,s')$ is conditioned on the whole transition, the learning of latent space can be achieved by gradient descent of $\nabla_{\varphi} \mathbb{E}_{q_{\varphi}(\mathbf{z}|s,a,s')}[\log p(s'|s,a,\mathbf{z})]$. As a result, the learning of useful latent variables is automatic in VDM training, and we do not need to select the latent variables manually.

\begin{figure}[!t]
\centering
\includegraphics[width=2.6in]{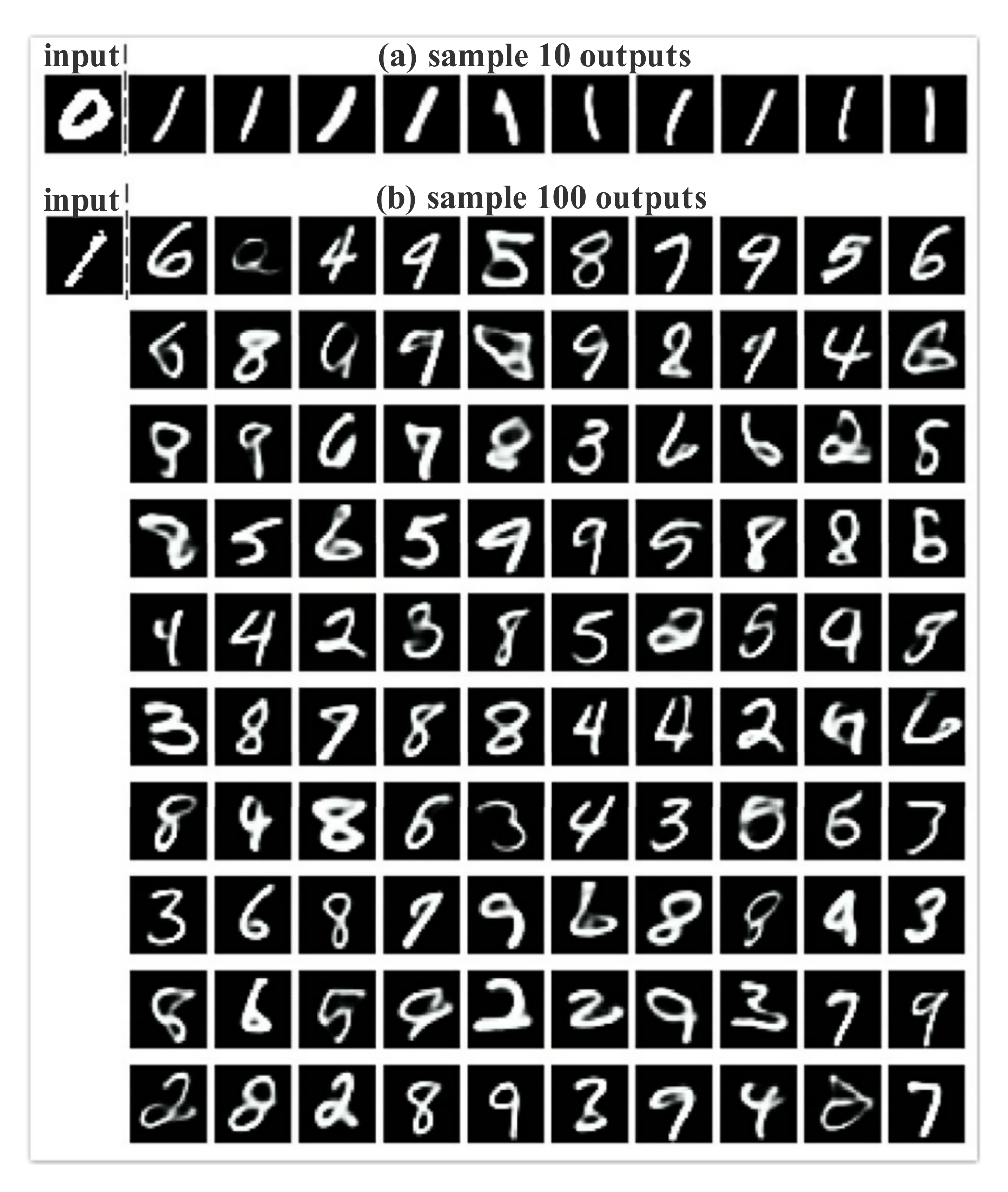}
\label{fig-VDM-res1}
\caption{Result of VDM in `Noisy-Mnist'. (a) When we input an image of digit `0', we sample 10 latent variables $\{\mathbf{z_1},...,\mathbf{z_{10}}\}$ and generate a next-state prediction for each one. VDM generates digit `1' with different writing styles. (b) When we input an image of digit `1', we sample 100 latent variables $\{\mathbf{z_1},...,\mathbf{z_{100}}\}$ and 100 next-state predictions are generated. The digits `2' to `9' are sampled with almost equal probability in various writing styles.}
\label{fig_VDM}
\vspace{-1em}
\end{figure}

As an example, we model the transition dynamics in MDP of `Noisy-Mnist' in Fig. \ref{fig_noise_mnist_mdp}. We first use an ensemble-based model that contains three individual encoder-decoder networks as a baseline. According to a resent research in model-based RL \cite{chua2018deep}, the ensemble model with probabilistic neural networks achieves the state-of-the-art performance in model-based planning. Each probabilistic network outputs a Gaussian distribution with diagonal covariances, and an ensemble of probabilistic networks captures the uncertainty in long-term dynamics-prediction. This probabilistic-ensemble model is incorporated into model predictive control (MPC) planning for policy search, and achieves strong performance in continuous control tasks \cite{chua2018deep}. Following this principle, we implement an ensemble-based dynamics model with three probabilistic neural networks for Noisy-Mnist. Each network outputs a 512d diagonal Gaussian to model the mean and variance of each pixel. We train such an ensemble-based model by supervised learning for 200 epochs to predict the next state. 

The ensemble-based baseline contains three individual encoder-decoder networks. As shown in Fig.~\ref{fig_ensemble}, three images are generated from each model with the same input. We do not average the outputs of the three models. In~(a), we use the image of digit `0' as the input and generate a prediction from each network in the ensemble model. The three images from three networks have correct image class but lack diversity in different writing styles. In~(b), we use the image of digit `1' as the input. We observe that each generated image from the ensemble model tends to average various reasonable next-state predictions with different classes and writing styles, which leads to blurred images. The ensemble model fails to learn the stochasticity and multimodality of the dynamic in this task. 

We analyze the possible reasons in the following. (\romannumeral1) The probabilistic-ensemble model proposed in \cite{chua2018deep} is used in continuous control tasks, where the state is low-dimensional and unstructured. However, Noisy-Mnist has high-dimensional image-based observations. The probabilistic ensemble may not suitable for this setting. (\romannumeral2) We focus on single-step dynamics modeling and use the prediction likelihood to encourage exploration, while the probabilistic ensemble in \cite{chua2018deep} focuses on long-term prediction to long-term planning through MPC. That is, the probabilistic-ensemble may not good at modeling the stochasticity in single-step transition, but it enables to capture the long-term uncertainty. (\romannumeral3) Noisy-Mnist is manually constructed and has high stochasticity in dynamics. In a practical RL environment, there may only exist finite modes and stochasticity. We use this task to illustrate that VDM works well even in extremely stochastic environments. We release the implementation of probabilistic-ensemble\footnote{https://github.com/Baichenjia/probabilistic-ensemble} for reproducibility and further improvement. In addition, applying VDM for long-term model-based planning is an interesting future direction.

We further perform VDM to learn the transition dynamic in `Noisy-Mnist' as a comparison. In this case, we train VDM in raw pixel space to facilitate visualization. We set the latent variable to be $\mathbf{z}\in \mathbb{R}^{64}$. We train VDM by maximizing $L_{\rm VDM}$ for 200 episodes. After training, we use the trained model to perform conditional inference to generate next-state predictions, as shown in Fig.~\ref{fig_VDM}. In~(a), we input an image of handwritten digit `0' as the current state $s$, and sample 10 latent variables $\{\mathbf{z_1},\ldots,\mathbf{z_{10}}\}$ from the prior $p_{\psi}(\mathbf{z}|s,a)$, where $a$ is trivial in this case and is set to be a constant vector. We feed each sampled latent variable into the generative network $p_{\theta}(s'|s,a,\mathbf{z})$, and we use the mean of $p_{\theta}$ as the next-state prediction. We observe that the generated images have the correct class with different writing styles. Similarly, in~(b), we input an image of handwritten digit `1' and sample 100 latent variables $\{\mathbf{z_1},\ldots,\mathbf{z_{100}}\}$ from the prior. Then 100 next-state predictions are generated. The images generated by different latent variables represent different classes of digits with various writing styles. We observe that digits from `2' to `9' are generated with almost equal probability, which is consistent with the transition dynamics of `Noisy-Mnist'. Moreover, each image generated is clear and does not mix up multiple possible outcomes. Thus, we conclude that VDM outperforms the ensemble-based model in modeling different modes and stochasticity of the `Noisy-Mnist'. 

\subsection{Intrinsic Reward for Exploration}\label{VDM-s4}

Given a transition tuple $(s,a,s')$, a desirable intrinsic reward that facilitates exploration takes the following form,
\begin{equation}\label{eq:222}
r^{i}:=-\log p(s'|s,a).
\end{equation}
Such a reward is desirable since a large reward corresponds to a small log-likelihood, which further indicates that the agent is unfamiliar with the corresponding transition. However, the reward $r^{i}$ defined in \eqref{eq:222} cannot be derived from VDM directly. To this end, we use an upper bound of $r^{i}$ as the intrinsic reward instead. We compute the intrinsic reward by first sampling several latent variables from the posterior and averaging the corresponding estimations of $L_{\rm VDM}$ from the sampled latent variable. We denote the intrinsic reward computed by sampling $k$ latent variables by $r^{i}_{k}$, which is defined as follows,
\begin{equation}
r^{i}_{k}:=-\mathbb{E}_{\mathbf{z}_i\sim q_{\varphi}(\mathbf{z}|s,a,s')}\Big[\log\frac{1}{k}\sum\nolimits_{i=1}^{k}w_i(\mathbf{z}_i)\Big],
\label{eq-reward}
\end{equation}
where $w_i(\mathbf{z}_i)$ is the estimation of the upper bound in \eqref{eq-jensen}, which takes the form of $w_i(\mathbf{z}_i)=p(s',\mathbf{z}_i|s,a)/q(\mathbf{z}_i|s,a,s')$. The following theorem shows that the reward $r^i_k$ defined in \eqref{eq-reward} is indeed an upper bound of $r^i$ defined in \eqref{eq:222}.

\newtheorem{theorem}{Theorem}
\begin{theorem}\label{thm:111}
It holds for all positive integers $m \leq k$ that
\begin{equation}
r^{i}\leq r^{i}_{k}\leq r^{i}_{m}.
\end{equation}
Moreover, if $w_i$ is bounded, then $r^{i}_{k}$ converges to $r^{i}$ as $k\to+\infty$.
\label{theorem1}
\end{theorem}

\begin{proof}
By \eqref{eq-jensen}, the following upper bound holds,
\begin{equation}
r^{i}_{k}=-\mathbb{E}\biggl[{\log}\frac{1}{k}\sum_{i=1}^{k}w_i(\mathbf{z}_i)\biggr]\geq -{\log}\mathbb{E}\biggl[\frac{1}{k}\sum_{i=1}^{k}w_i(\mathbf{z}_i)\biggr]=r^{i}.
\end{equation}

Let $I\in\{1,...k\}$ with $|I|=m$ be a uniformly distributed subset of distinct indices from $\{1,...,k\}$. It then holds that $\mathbb{E}_{I=\{i_1,...,i_m\}}\Bigl[\frac{a_{i_1}+...+a_{i_m}}{m}\Bigr]=\frac{a_1+...+a_k}{k}$ for any sequence of numbers $a_1,...,a_k$. Thus, following from Jensen's inequality, we have
\begin{equation}
\begin{split}
r^{i}_{k} &= -\mathbb{E}_{\mathbf{z}_1,...,\mathbf{z}_k}\biggl[{\log}\frac{1}{k}\sum_{i=1}^{k}w_i(\mathbf{z}_i)\biggr]\\ 
&= -\mathbb{E}_{\mathbf{z}_1,...,\mathbf{z}_k}\biggl[{\log}\mathbb{E}_{I=\{i_1,...,i_m\}}\Bigl[\frac{1}{m}\sum_{j=1}^{m}w_j(\mathbf{z}_j)\Bigr]\biggr]\\
&\leq-\mathbb{E}_{\mathbf{z}_1,...,\mathbf{z}_k}\biggl[\mathbb{E}_{I=\{i_1,...,i_m\}}\Bigl[{\log}\frac{1}{m}\sum_{j=1}^{m}w_j(\mathbf{z}_j)\Bigr]\biggr]\\ 
&=-\mathbb{E}_{\mathbf{z}_1,...,\mathbf{z}_m}\biggl[{\log}\frac{1}{m}\sum_{i=1}^{m}w_i(\mathbf{z}_i)\biggr]=r^{i}_{m}.
\end{split}
\end{equation}

Moreover, we consider the random variable $M_k=\frac{1}{k}\sum_{i=1}^{k}w_i(\mathbf{z}_i)$. If $w_i(\mathbf{z}_i)$ is bounded, then it follows from the strong law of large numbers that $M_k$ converges to $-\mathbb{E}_\mathbf{z}[w_i]$ almost surely. Hence $r^{i}_{k}=-\mathbb{E}{\rm log}[M_k] \to -\log\mathbb{E}_\mathbf{z}[w_i] = r_i$ as $k\rightarrow\infty$.
\end{proof}

Theorem \ref{thm:111} suggests that, by sampling more latent variables to calculate the intrinsic reward $r^i_k$, we obtain a tighter upper-bound of $r^i$ as the intrinsic reward for self-supervised exploration. The agent trained to maximize the intrinsic reward will favor transitions with low log-likelihood. The intrinsic rewards are high in areas where less explored by the agent, or in areas with complex dynamics that contain multimodality and stochasticity.

The complete procedure of self-supervised exploration with VDM is summarized in Algorithm~\ref{alg1}. In each episode, the agent interacts with the environment to collect the transition $s_t,a_t,s_{t+1}$. Then a random CNN that extracts feature is utilized to embed the states, and the embedded transition is fed into VDM to estimate the intrinsic reward following \eqref{eq-reward}. After the end of an episode, we use the collected $T$ transitions to update the parameters of the policy by following the PPO gradient defined in \eqref{eq-ppo1} and \eqref{eq-ppo2} associated with generalized advantage estimation~\cite{gae-2016}. Meanwhile, we update the proposal network, prior network and generative network in VDM based on \eqref{eq-reparameter} and \eqref{eq-second-kl}. We further use the updated policy and VDM for the interaction in the next episode. 

\begin{algorithm}[t]
\caption{Self-supervised exploration with VDM}
\label{alg1}
\begin{algorithmic}[1]
\STATE {\bf Initialize:} The actor-critic network of PPO, the VDM networks $(\varphi,\psi,\theta)$, and a random CNN $\phi$
\FOR {episode $i = 1$ {\bfseries to} $M$}
\FOR {timestep $t = 0$ {\bfseries to} $T-1$}
\STATE Getting action from actor $a_t=\pi(s_t)$, execute the action and observe the next state $s_{t+1}$.
\STATE Putting $\{s_t,a_t,s_{t+1}\}$ in VDM to calculate the intrinsic reward $r^i_{k(t)}$ by following~\eqref{eq-reward}.
\ENDFOR
\STATE Using $\{s_t,a_t,r^i_{k(t)},s_{t+1}\}_{t=0}^{T-1}$ to update the actor-critic network with PPO.
\STATE Using $\{s_t,a_t,s_{t+1}\}_{t=0}^{T-1}$ to calculate the objective function $L_{\rm VDM}$ by following~\eqref{eq-lb}.
\STATE Taking stochastic gradient ascent $t_{\rm vdm}$ times to maximize $L_{\rm VDM}$ and update parameters $(\varphi,\psi,\theta)$ of VDM.
\ENDFOR
\end{algorithmic}
\end{algorithm}

\section{Experiments}

In this section, we conduct experiments to compare the proposed VDM with several state-of-the-art model-based self-supervised exploration approaches. We first describe the experimental setup and implementation detail. Then, we compare the proposed method with baselines in several challenging image-based RL tasks. The code and video are available at {\url{https://sites.google.com/view/exploration-vdm}.}

\subsection{Environments}

We evaluate the proposed method on several challenging image-based tasks from OpenAI Gym\footnote{\url{http://gym.openai.com/}} and Retro\footnote{\url{https://retro.readthedocs.io}}, including

 (\romannumeral1) \textbf{Atari games}. The standard Atari games have high-dimensional image-based observations. Although the dynamic is known to be near-deterministic, it is stochastic given only a limited horizon of past observed frames. 

(\romannumeral2) \textbf{Atari games with sticky actions}. We add stochasticity in the environment of Atari games by making actions `sticky' \cite{jalr-2018}. More specifically, in each time step, the previously executed action is repeated with a probability of $\tau$. 

(\romannumeral3) \textbf{Super Mario}. There are several levels in Super Mario games. We change different levels upon training the model to evaluate the adaptability of VDM to the novel environments. 

(\romannumeral4) \textbf{Two-players pong}. Both sides of the players are controlled by the intrinsic-driven agents that fight against each other. The stochasticity of the environment comes from the opponent, whose policy evolves along with the training process. 

(\romannumeral5) \textbf{Real robotic manipulating task}. Finally, we evaluate our method in real-world robotic manipulation scenarios.

\subsection{Experimental baselines}

To validate the effectiveness of our method, we compare the proposed method with the following self-supervised exploration baselines. Specifically, we conduct experiments to compare the following methods: (\romannumeral1) \textbf{VDM}. The proposed self-supervised exploration method. (\romannumeral2) \textbf{ICM}~\cite{curiosity-2017}. ICM first builds an inverse dynamics model to contain information related to the actions taken by the agent while ignoring other side information. ICM utilizes the prediction error of the forward dynamics as the intrinsic reward. ICM is robust to the stochasticity of the environment. (\romannumeral3) \textbf{RFM}~\cite{largescale-2019}. Similar to ICM, RFM uses the prediction error as the intrinsic reward for self-supervised exploration. RFM uses a fixed CNN to extract features of the state. RFM achieves comparable performance with ICM in challenging tasks and is more computationally efficient. (\romannumeral4) \textbf{Disagreement}~\cite{disagree-2019}. The disagreement method uses an ensemble of environment models to evaluate the uncertainty in exploration. The agent is encouraged to explore the area with maximum disagreement among the predictions in the ensemble of models. Such a disagreement can be cast as Bayesian measures of model uncertainty~\cite{sekar2020planning} and prevents the agent from getting stuck in local-minima of exploration.

We compare the model complexity of all the methods in Table \ref{tab:compare-para}. VDM, RFM, and Disagreement use a fixed CNN for feature extraction. Thus, the trainable parameters of feature extractor are 0. ICM estimates the inverse dynamics for feature extraction with 2.21M parameters. ICM and RFM use the same architecture for dynamics estimation with 2.65M parameters. Disagreement utilizes a dynamics ensemble that contains 26.47M parameters. VDM only requires slightly more parameters (2.73M) than an ordinary dynamics model (in ICM \& RFM), since the latent space is relatively low dimensional with $128$ dimensions. 
\begin{table}[t]
\centering
\caption{Comparison of model complexity}
\begin{tabular}{c|rr|r}
    \hline
    {~} & Feature extractor & Dynamic model & \textbf{Total}\cr
    \hline
    ICM          & 2.21M  & 2.65M &  4.86M \\
    RFM          & 0M     & 2.65M &  2.65M \\
    Disagreement & 0M     & 26.47M &  26.47M \\
    \textbf{VDM (ours)}   & 0M     & 2.73M &  2.73M \\    
\hline
\end{tabular}\label{tab:compare-para}
\end{table}

\subsection{Implementation details}

We first introduce the common implementation over our methods and all baselines, then specify the details of the proposed VDM used in our method. Tab.~\ref{tab-parameter} summaries the hyper-parameters of VDM.

\begin{table}[t]
  \caption{Summary of Hyper-parameters in VDM}
  \label{tab:hyper-atari}
  \centering
  \begin{tabular}{p{0.15\columnwidth}p{0.15\columnwidth}p{0.55\columnwidth}}
    \hline
    Type & Module & Hyper-parameter details \\
    \hline
    \multirow{6}*{\makecell[c]{Common \\ \textbf{(not tuned)}}} & Preprocessing & \ding{192} Resize; \ding{193} Grayscale; \ding{194} Framskip \\
	& Optimization & \ding{192} Network; \ding{193} Baselines for variance reduction; \ding{194} Learning rate; \ding{195} Adam; \ding{196} Entropy regularizer; \ding{197} $\gamma$; \ding{198} $\lambda$; \ding{199} Epochs \\
    & Normalization & Normalize advantage and reward \\
    & Multi-actor & 128 parallel actors \\ 
    \hline 
    \multirow{2}*{\makecell[c]{VDM-specific \\ \textbf{(tuned)}}}  & Architecture & Posterior, prior and generative networks \\
    & Optimization & \ding{192} Learning rate; \ding{193} Training speed; \ding{194} $k$\\
    \hline
  \end{tabular}\label{tab-parameter}
\vspace{-1em}
\end{table}

\textbf{Common implementation.}
All baselines are based on the same basic implementation. The key elements including state preprocessing, PPO algorithm, normalization methods, and multiple actors. 

\emph{State preprocessing}. 
In Atari games, the observations are raw images. The images are resized to $84\times 84$ pixels and converted to grayscale. The state stacks $4$ recent observations as a frame of shape $84\times 84\times 4$. In both Super Mario and Atari games, we use the frame-skip technique to repeat each action four times. Before training, the agent interacts with the environments for $10^4$ steps to estimate the mean and standard deviation of the states. We further normalize the observed states for training by the estimated mean and standard deviation before training. 

\begin{figure*}[!t]
\centering
\includegraphics[width=7.0in]{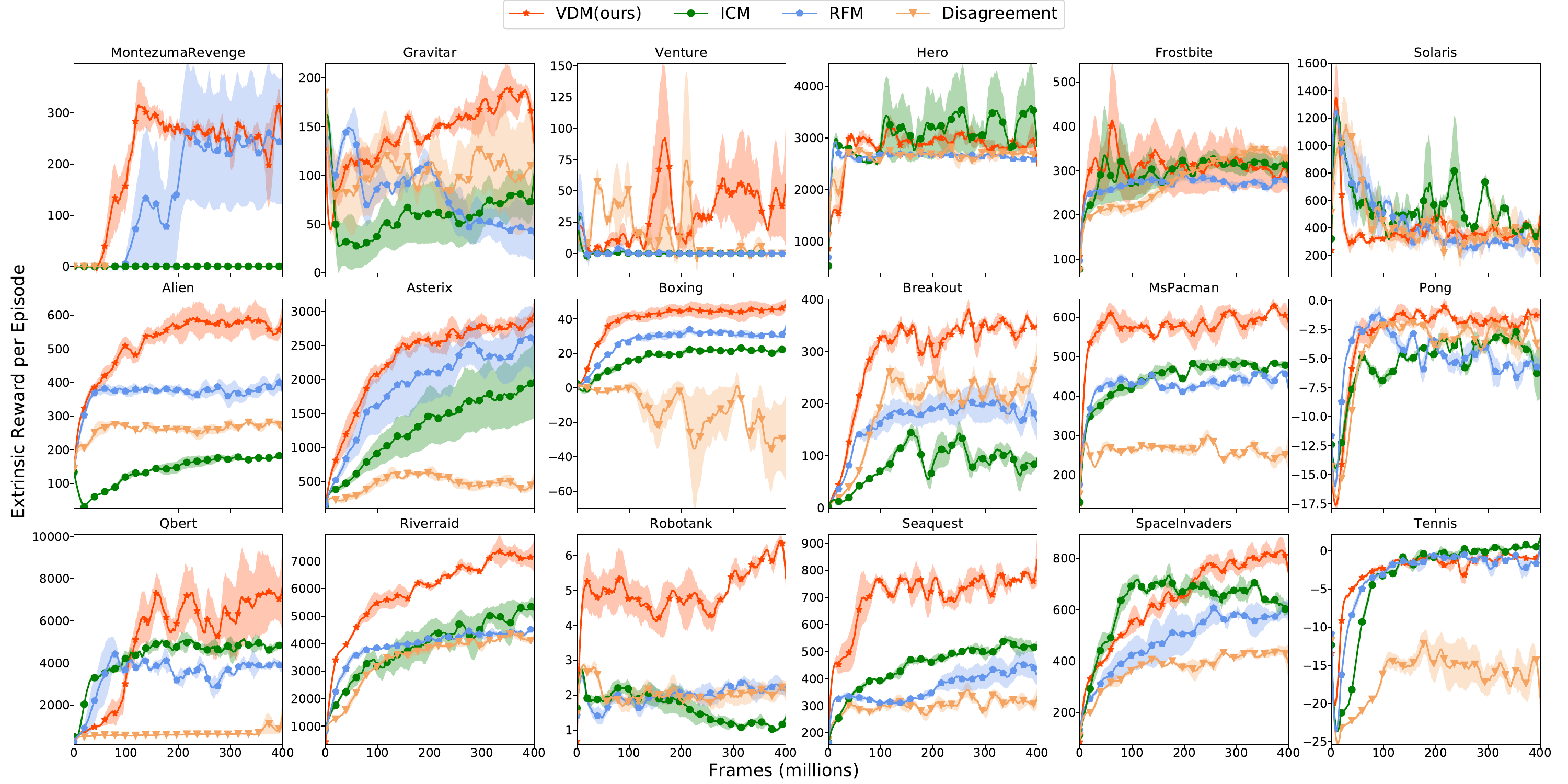}
\caption{The evaluation curve in Atari games. The first 6 games are hard exploration tasks. The different methods are trained with different intrinsic rewards, and extrinsic rewards are used to measure the performance. Our method performs best in most games, both in learning speed and quality of the final policy. The agent aims at staying alive and exploring the complex areas by maximizing the intrinsic rewards from VDM.}
\label{fig-result-atari}
\vspace{-1em}
\end{figure*}

\emph{PPO algorithm}. The actor-network in PPO contains $3$ convolution layers and $2$ fully-connected layers. The filters of each convolution layer are $32$, $64$, and $64$, respectively. The corresponding kernel-sizes are $8$, $4$, and $3$, respectively. The state is embedded into a $512$-dimensional feature vector after passing the convolution network. We further feed the embedded state into two fully-connected layers, each with $512$ hidden units. The outputs of the policy and the value-function are split in the last layer. We use the Softmax distribution as the output of policy in Atari and Super Mario. In the robotic manipulating task, the output of policy is the mean and variance of the diagonal Gaussian distribution. The output of the value function is a single unit that indicates the value estimation of the current state. PPO also includes an entropy regularizer to prevent premature convergence with a ratio of $10^{-3}$. We use Adam optimizer to update the parameters. We set the learning rate at $10^{-4}$ and hyperparameters $\gamma$ and $\lambda$ of PPO at $0.99$ and $0.95$.

\emph{Normalization methods}. We normalize the intrinsic reward and advantage function in training for more stable performance. Since the reward generated by the environment are typically non-stationary, such normalization is useful for a smooth and stable update of the value function. In practice, we normalize the advantage estimations in a batch to achieve a mean of $0$ and a standard deviation of $1$. We smoothen the intrinsic rewards exponentially by dividing a running estimate of the standard deviation of the sum of discounted rewards. 

\emph{Multiple actors}. We use MPI to perform 128 parallel actors in all baselines and environments to gather experiences. 
%Using multiple actors increase the batch size in training and increase the stability of the policy learned.

\textbf{VDM-specific implementation.} VDM consists of a proposal network, a prior network, and a generative network. In addition, we feed the states into a random CNN to obtain a $512$-dimensional feature vector for the training of VDM. 

\emph{Network architecture}. The proposal network contains $2$ fully-connected layers and $3$ residual blocks. The input to the proposal network contains features of the current state, next state, and action. In each layer, we integrate the action with features from the previous layer, which amplifies the impact of actions. Each residual block contains two dense layers. We use the skip-connection for the input and output in the residual blocks. The output of the proposal network is a $256$-dimensional vector, which contains the mean and standard deviation of the Gaussian latent variable $\mathbf{z}$. In addition, we use the soft-plus activation function to keep the standard deviation positive. The prior network has a similar architecture as the proposal network. The generative network uses the latent variable as input. We sample the $128$-dimensional latent variable based on the output of the proposal network, and feed the latent variable into $3$ dense layers and $3$ residual blocks to get the $1024$-dimensional output, which consists of the mean and standard deviation of the next-state prediction. We further construct the skip-connection between the prior network and the generative network. 

\emph{Optimization detail}. We update the parameters of VDM for $t_{\rm vdm}$ times after each episode by using Adam optimizer with the learning rate of $10^{-4}$. The hyper-parameter $t_{\rm vdm}$ controls the training speed of VDM. We set $t_{\rm vdm}=3$ in all the tasks. The intrinsic reward $r^i_k$ defined in \eqref{eq-reward} contains a hyper-parameter $k$, which is the number of latent variables sampled for reward estimation. We set $k=10$ in all the tasks. We refer the ablation study of hyper-parameters in \S\ref{self-compare}.
%We highlight that if $t_{\rm vdm}$ is too large, the VDM may overfit to the current scenarios. In contrast, if $t_{\rm vdm}$ is too small, the intrinsic reward decreases slowly in the same scenarios during the exploration. 

\subsection{Result comparison.}

\subsubsection{Atari games}

We first evaluate our method on standard Atari games. Since different methods utilize different intrinsic rewards, the intrinsic rewards are not applicable to measure the performance of the trained purely exploratory agents. In alternative, we follow~\cite{largescale-2019,disagree-2019}, and use the extrinsic rewards given by the environment to measure the performance. We highlight that the extrinsic rewards are \emph{only} used for evaluation, not for training. We illustrate the evaluation curves of $18$ common Atari games in Fig.~\ref{fig-result-atari}, where the first $6$ games are hard exploration tasks. We draw each curve with five distinct random seeds. For each method, the solid line indicates the mean episodic reward of all five seeds, and the shadow area shows the confidence interval (i.e., $\pm$Std of episodic rewards among all seeds) of the performance. The result shows that self-supervised exploration enables the agent to obtain higher extrinsic rewards by learning based on intrinsic rewards. More specifically, maximizing the intrinsic rewards encourages the agent to explore the complicated part of the environment, which typically corresponds to significant changes in the scenarios and leads to large extrinsic rewards. 
%Previous work~\cite{onlearning-2018} has pointed out that the optimal intrinsic reward function should be one that maximizes the task-specifying extrinsic rewards. The consistency of intrinsic and extrinsic rewards verifies this idea. 

\begin{figure*}[!t]
\centering
\includegraphics[width=7.0in]{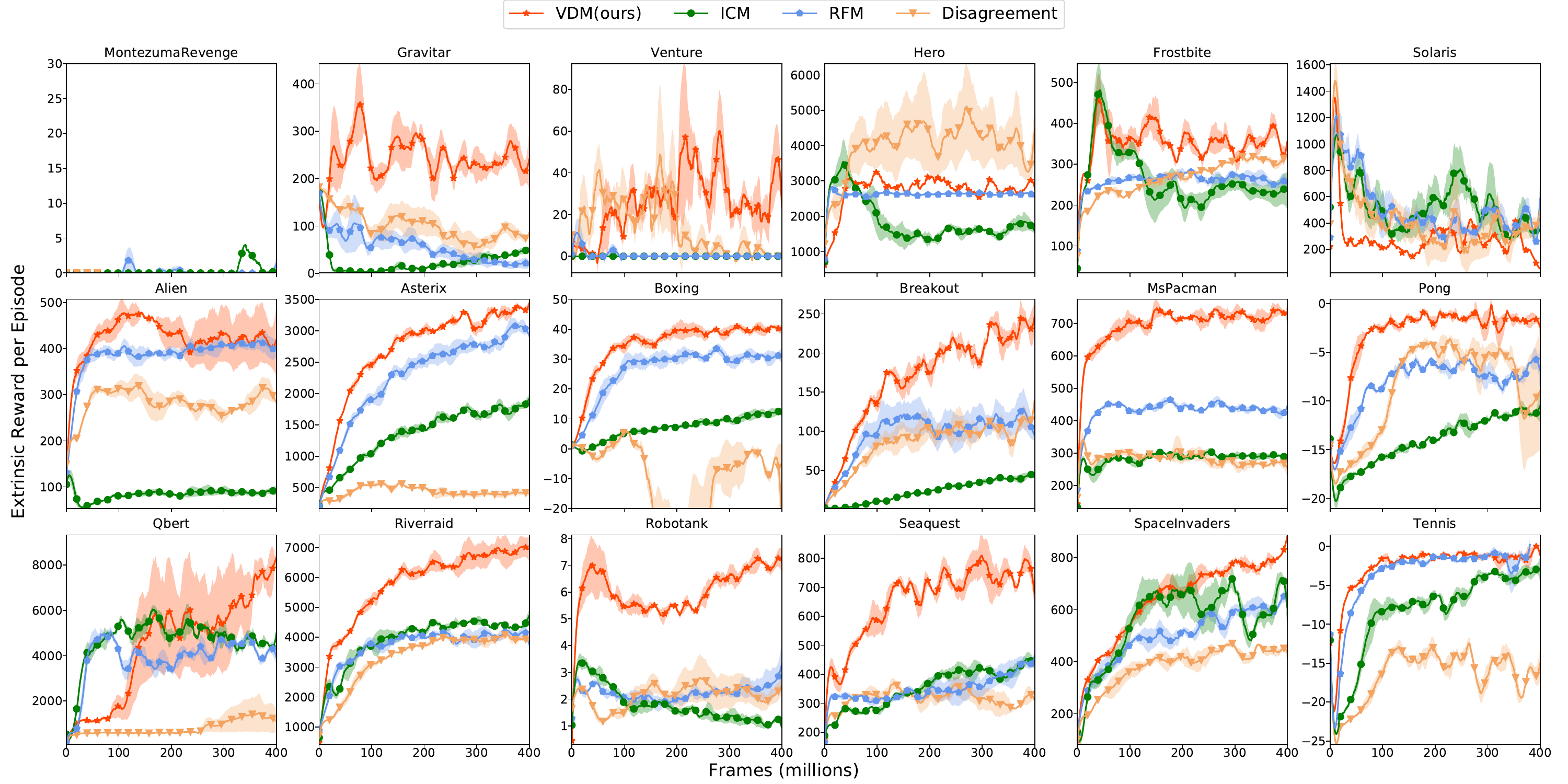}
\caption{The evaluation curve in Atari games with sticky actions. Such an environment introduces additional stochasticity in Atari games. As shown in figures, our proposed method outperform all the baselines in most games. We highlight that the VDM based variational inference uses the latent variable to model variation in modes and stochasticity in transition dynamics, which makes the proposed method suitable for exploration in complex environments.}
\label{fig-result-atari-sticky}
\end{figure*}

\begin{figure*}[!t]
\centering
\subfigure[Mario Levels]{\includegraphics[width=2.1in]{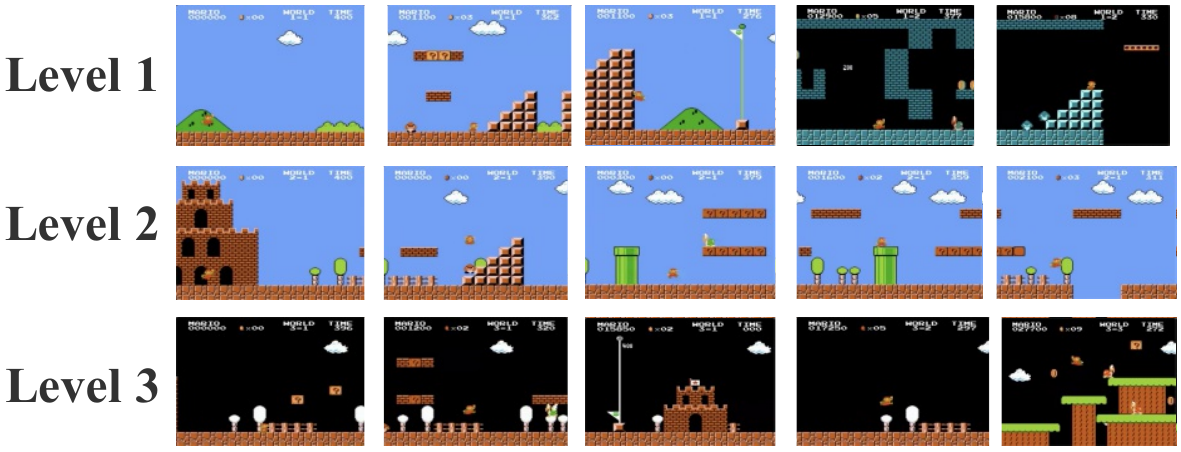}\label{fig-Mario-all-levels}}
\hspace{0.2em}
\subfigure[Results of Level 1]{\includegraphics[width=1.5in]{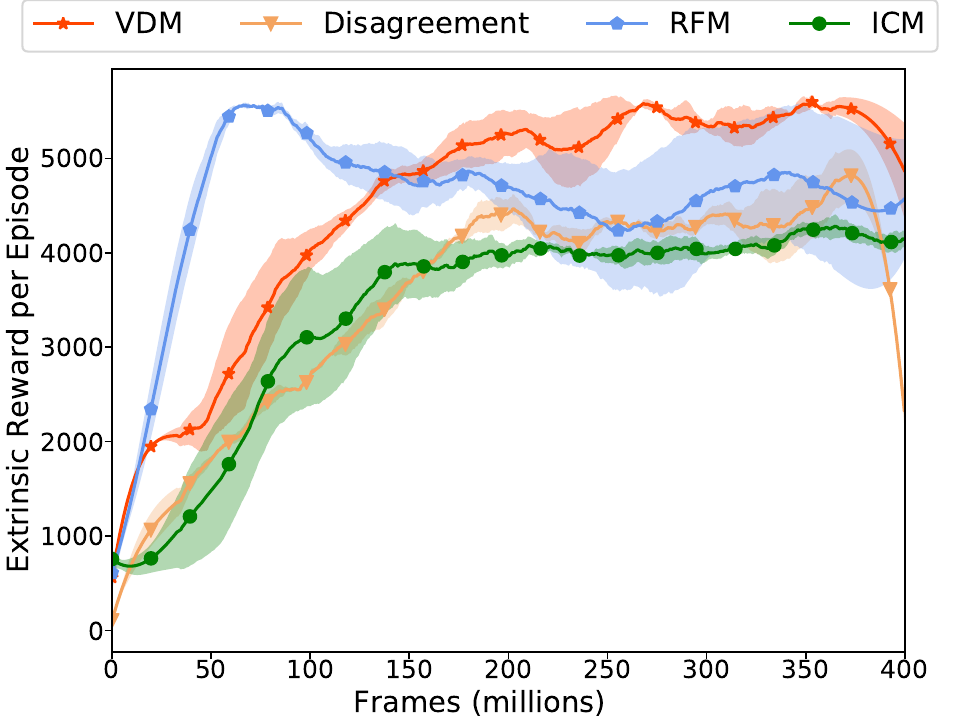}\label{fig-Mario-level-1}}
\hspace{0.1em}
\subfigure[Transfer in Level 2]{\includegraphics[width=1.5in]{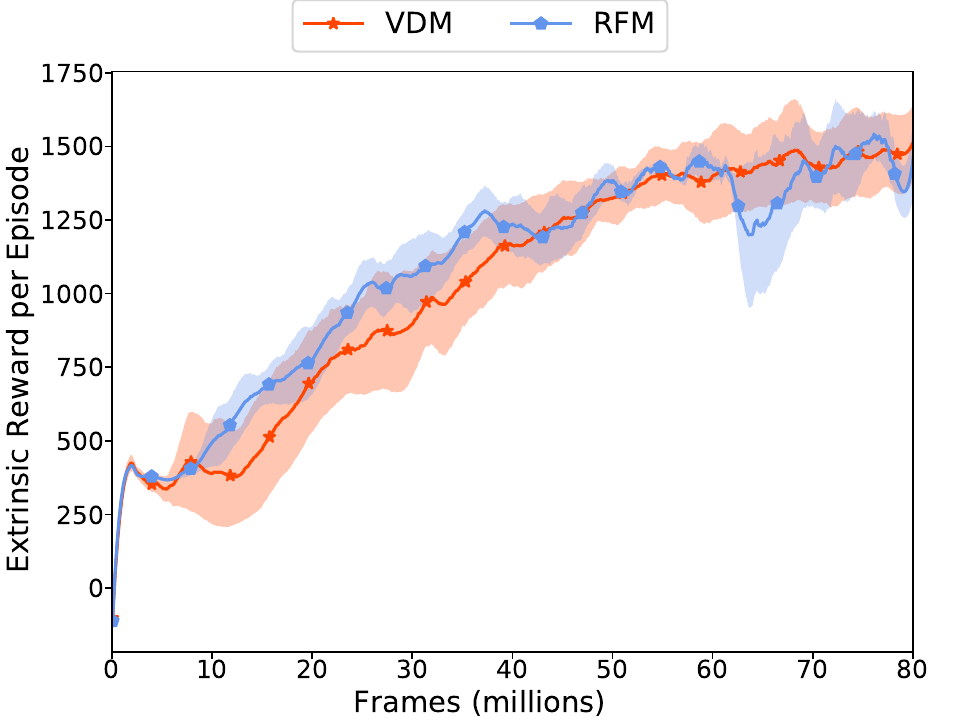}\label{fig-Mario-level-2}} 
\hspace{0.1em}
\subfigure[Transfer in level 3]{\includegraphics[width=1.5in]{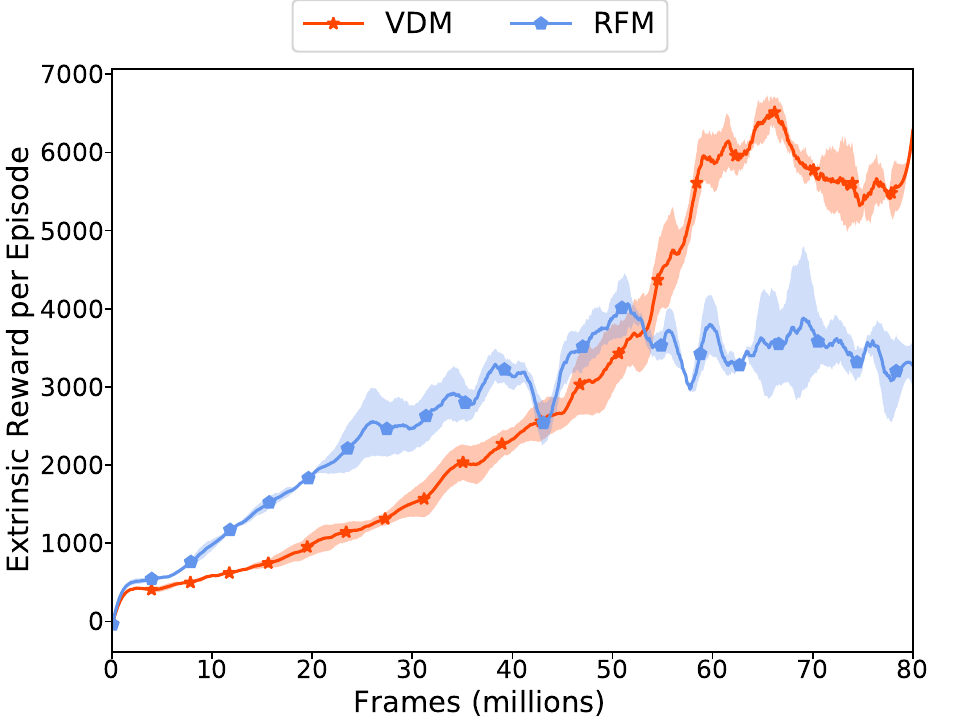}\label{fig-Mario-level-3}}
\caption{Result comparison in Super Mario. (a) We show 5 screenshots at each level. (b) Training from scratch in level 1. (c) Transfer from level 1 to level 2. (d) Transfer from level 1 to level~3.}
\label{fig-mario-res}
\vspace{-1em}
\end{figure*}

We observe that our method performs the best in most of the games, in both the sample efficiency and the performance of the best policy. The reason our method outperforms other baselines is the multimodality in dynamics that the Atari games usually have. Such multimodality is typically caused by other objects that are beyond the agent's control. Affected by these objects, taking the same action may yield different outcomes. For example, in \emph{MsPacman}, the ghosts choose directions at each fork of the maze freely, which is beyond the control of the agent. Similar to different image classes in the Noisy-Mnist example, different behavior of ghosts leads to the different modes in the transition dynamics. VDM captures the multimodality of the dynamic when measuring the novelty of transitions, which leads to better intrinsic rewards for exploration. Moreover, in VDM, the features encoding multimodality and stochasticity are contained in posterior and prior networks separated from the reconstruction features in the generative network. Hence, VDM prevents the features of multimodality and stochasticity from being ruined in the training of the generative model.

Nevertheless, the introduce of latent variable often introduce instability to neural networks. For example, the popular deep learning models like VAEs and GANs are shown to be unstable since the introduce of stochasticity in latent space \cite{stable-2,stable-3}. We find VDM performs generally well and shows small performance variance in most of the tasks. However, there exists several games, such as Qbert and Venture, where policies trained with VDM have relatively high variance in some seeds. We believe the occurrence of outliers are acceptable as VDM shows stable and improved performance in most of the tasks.

\subsubsection{Atari games with sticky actions}

To evaluate the robustness of exploration methods, we conduct experiments on sticky Atari games, which introduce stochasticity in Atari games. Following \cite{jalr-2018}, we use a parameter $\tau$ to control the stickiness in Atari games. Specifically, in time step $t$, the environment repeats the agent’s previous action $a_{t-1}$ with probability $\tau$, while executing the action $a_t$ selected by the agent with probability $1 - \tau$. We use $\tau=0.25$ in our experiments. Learning in sticky Atari games is more challenging due to the additional stochasticity in transitions. For example, in \emph{Qbert}, the sticky action increases the chance of falling off the pyramid when the agent lands on edge, even if the agent switches to a no-op action. Thus, efficient exploration methods need to consider such stochasticity in the transition dynamics. As shown in Fig.~\ref{fig-result-atari-sticky}, the performance of most baselines in sticky Atari is greatly undermined compared with the standard Atari games, while VDM is scarcely affected by sticky actions. Besides, the disagreement~\cite{disagree-2019} based on Bayesian uncertainty also demonstrates robustness in coping with sticky actions. In conclusion, our method outperforms all baselines in most games among the sticky Atari games. We highlight that VDM uses the latent variable to model variation in modes and stochasticity of the transition dynamics, making the proposed intrinsic rewards suitable for exploration in stochastic dynamics.

\subsubsection{Adaptability evaluation in Super Mario}

One reason to perform self-supervised exploration is to adapt the trained explorative agent in similar environments for exploration. To evaluate such adaptability, we conduct experiments on Super Mario. Super Mario has several levels of different scenarios. We take $5$ screenshots at each level when playing games, as shown in Fig. \ref{fig-Mario-all-levels}. Similar to \cite{largescale-2019}, we adopt an efficient version of Super Mario in Retro that simulates fast. The Level $1$ of the game has different scenarios of day and night. We train all methods from scratch in the Level $1$. We refer to Fig. \ref{fig-Mario-level-1} for the evaluation curves of extrinsic rewards. The proposed VDM shows similar performance to RFM, while VDM's learning curve is smoother and more stable.

To evaluate the adaptability, we further adopt the policies learned from the Level $1$ to other levels. More specifically, for each method, we first save the last policy when training in the Level $1$, and then fine-tune such a policy in the Levels $2$ and $3$. Since the VDM and RFM methods perform the best in the Level $1$, we conduct adaptability experiments exclusive on VDM and RFM. We illustrate the result in Fig.~\ref{fig-Mario-level-2} and Fig~\ref{fig-Mario-level-3}. We observe that VDM performs similarly to RFM at the Level $2$, while performing much better than RFM when transferring to the Level $3$. As a result, we conclude that the policy learned by VDM demonstrates better adaptability to novel environments.

\begin{figure}[!t]
\centering
\includegraphics[width=2.0in]{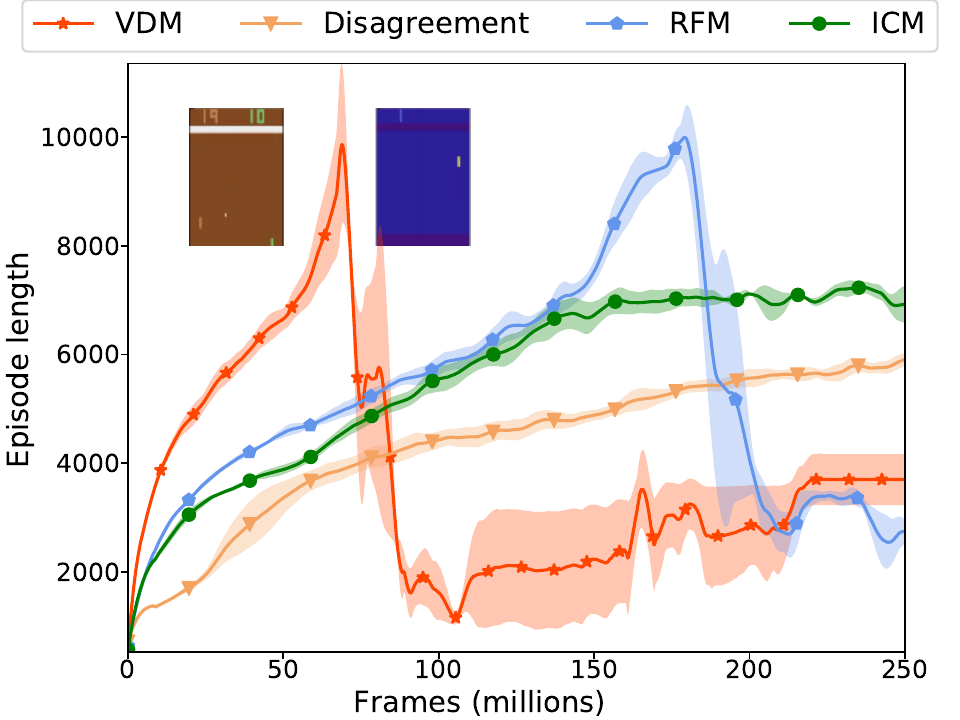}
\caption{Result comparison of evaluation curve in Two-players pong. The episode length become longer along with training time.}
\label{fig-multiplayer}
\vspace{-0.5em}
\end{figure}

\subsubsection{Two-players pong}

To further investigate the capability of our method in coping with highly stochastic environments, we conduct experiments on games where both the agent and its opponent are controlled by self-supervised exploratory policies. The stochasticity of the transition dynamics is much higher for both sides of the game since the opponent's evolution in policy changes the agent's transition. In contrast, when playing Atari games, the opponent is controlled by a hardcoded policy, which yields a relatively stable transition. We use the Two-player Pong game for the experiment. The extrinsic reward is not appropriate for evaluating different methods in this experiment, since both sides are controlled by policies that evolved together. Alternatively, we use the length of the episode as the evaluation criterion. Such a criterion is appropriate since, on the one hand, to maximize the intrinsic rewards, both the agent and its opponent aims to beat each other while avoiding the dead ball that terminates the episode. On the other hand, as the policy evolves, both sides of the game eventually utilize approximately the same policy, which yields a long episode.

We illustrate the results in Fig. \ref{fig-multiplayer}. We observe that the episode length becomes longer over training time with the intrinsic reward estimated from VDM, as anticipated. We observe that our method reaches the episode length of $10^4$ with the minimum iteration steps. After reaching the maximum episode length, the game rallies eventually get so long that they break our Atari emulator, causing the colors to change radically, which crashes the policy. The two images of observation in Fig. \ref{fig-multiplayer} illustrate the change of emulator. The RFM achieves similar results with twice the training step as of VDM. In conclusion, the pure-exploratory policy learned by VDM enables the control of both sides to improve their policies, which achieves the Nash equilibrium more efficiently in the Two-player Pong game than the baseline methods.

\subsubsection{Real robotic manipulating task}

\begin{figure}[!t]
\centering
\includegraphics[width=2.5in]{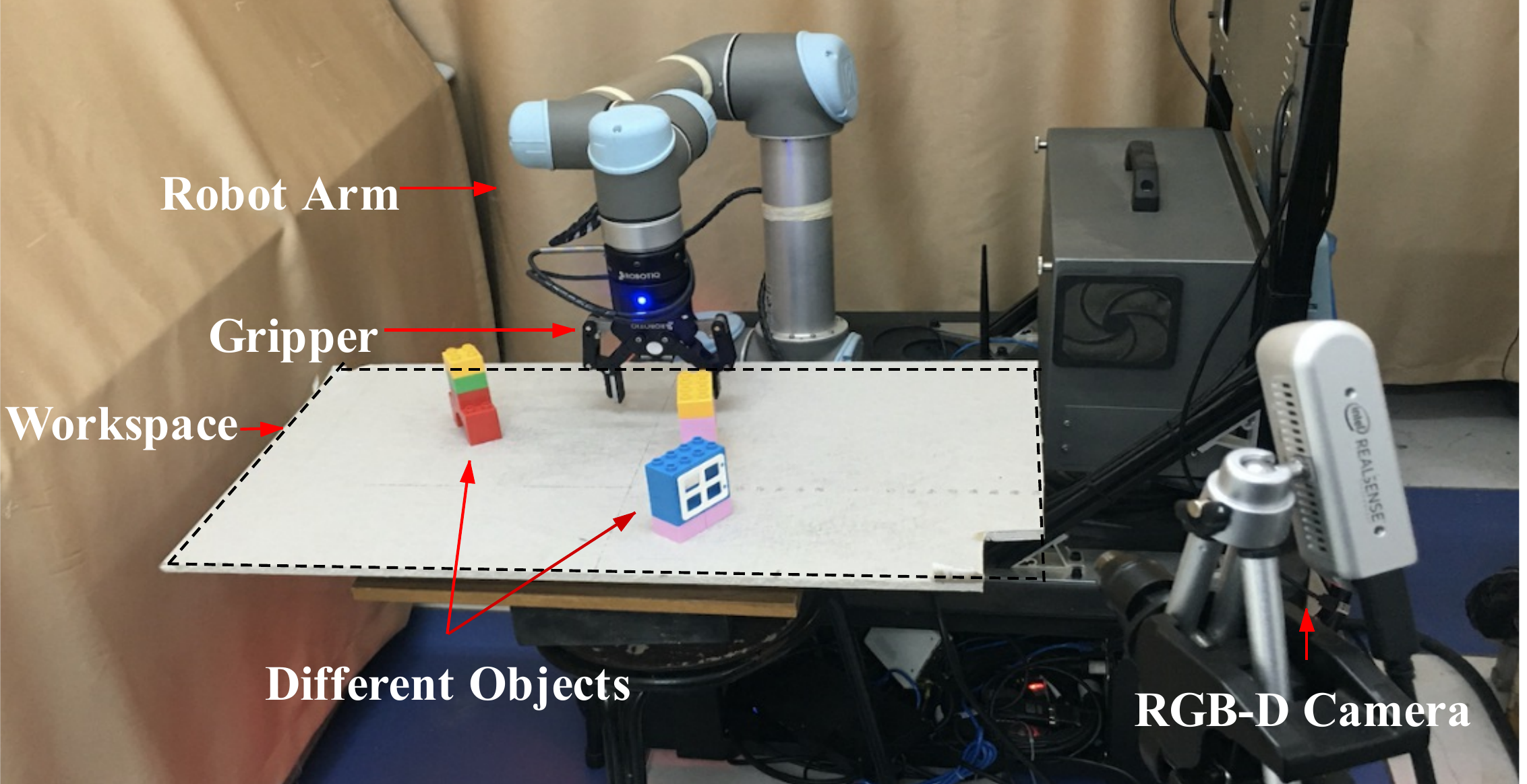}
\caption{The environment in real-world robotics experiment. The equipment mainly includes an RGB-D camera, a UR5 robot arm, and several objects.}
\label{fig-robot}
\vspace{-1em}
\end{figure}

\begin{figure}[!t]
\centering
\includegraphics[width=3.2in]{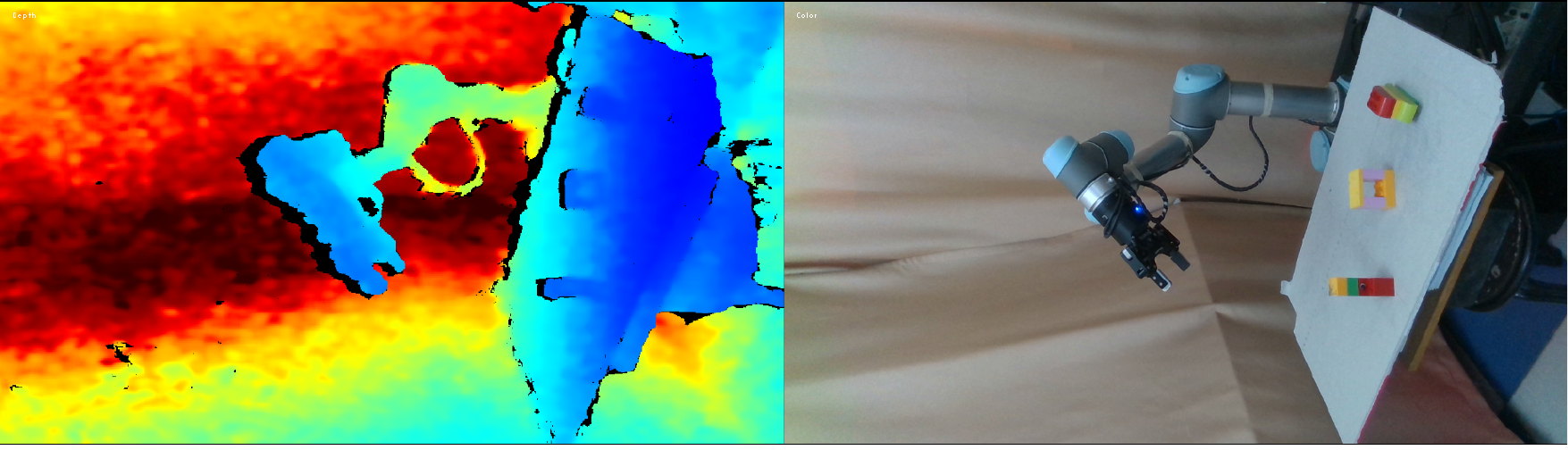}
\caption{An example of RGB-D data collected by the camera in training. The depth data shows on the left.}
\label{fig-robot-depth}
\vspace{-1em}
\end{figure}

Finally, to evaluate our proposed method in real-world tasks, we conduct experiments on the real-world robot arm to train a self-supervised exploration policy. We highlight that policy learning in a real robot arm needs to consider both the stochasticity in the robot system and the different dynamics corresponding to different objects. 

We demonstrate the setup of the experiment in Fig.~\ref{fig-robot}. The equipment mainly includes an RGB-D camera that provides the image-based observations, a UR5 robot arm that interacts with the environment, and different objects in front of the robot arm. An example of the RGB-D image is shown in Fig.~\ref{fig-robot-depth}. We develop a robot environment based on OpenAI Gym to provide the interface for the RL algorithm. We connect a GPU workstation, the robot arm, and a camera through TCP protocol. The PPO algorithm and VDM are running on the GPU workstation. During the training, the samples collected by the camera are sent to the GPU workstation, and the policy commands generated by the policy are sent to the robot arm to execute. We stack the RGB-D data and resize it to $84\times84\times4$ pixels as the input state in RL. The arm moves according to the position control of a vertically-oriented gripper. We represent the continuous actions by a Cartesian displacement $[dx,dy,dz,d\omega]$, where $\omega$ is the rotation of the wrist around the z-axis. The output of the policy is a Gaussian distribution. We do no use gripper in our experiment and keep the gripper open in training. Each training episode contains a maximum of $100$ time steps of interaction. An episode terminates when the experiment exceeds the maximal length of $100$ time steps, or the robot arm pushes all objects out of the workspace. 

\begin{table}[!t]
\centering
\caption{Evaluation of OIF in robot arm}
\label{table_robot}
\begin{tabular}{cccc}
	\hline
    {~}&\multicolumn{3}{c}{Training episodes}\cr\cline{2-4}
    {Method}&100&300&500\cr
    \hline
    VDM&24\%&40\%&55\% \\
    Disagreement&12\%&31\%&43\% \\
\hline
\end{tabular}
\vspace{-1em}
\end{table}

At the beginning of each episode, we put three objects in the workspace. Using fewer objects makes the robot arm harder to interact with the objects by taking actions randomly. We use a set of 10 different objects for training and $5$ objects for testing. We follow \cite{disagree-2019} and use the Object-Interaction Frequency (OIF) to evaluate the quality of the exploratory policy as follows. In a testing episode, We run the policy for $t_1$ time steps and record the number of steps $t_2$ that the robot interacts (i.e., touch) with the object. OIF is computed by ${\rm OIF}=t_2/t_1$. Maximizing the intrinsic reward estimated based on VDM encourages the agent to explore the critical part in the workspace, where the agent learns to interact with different objects. During the evaluation period, we reset the environment for every $10$ robot steps. We show the evaluation result in Table \ref{table_robot}. We compare our methods with Disagreement, which outperforms ICM and RFM in the real robotics system \cite{disagree-2019}. In addition, we include a random policy for exploration as a baseline. Our final policy interacts approximately $55$\% of times with unseen objects, whereas Disagreement achieves $43$\%, and the random policy achieves $2$\%. The video of the robot arm experiment is available at {\url{https://sites.google.com/view/exploration-vdm}.

\subsection{Ablation study}\label{self-compare}

In this section, we present the results of the ablation study of VDM. Recall that we have "Common" modules and "VDM-specific" models according to Tab.~\ref{tab-parameter}. Common modules are used for policy optimization rather than exploration, and all compared methods use the same common modules and not be tuned. "VDM-specific" modules include hyper-parameters for the proposed VDM, and these hyper-parameters are tuned through grid search \cite{exp-1,exp-2,exp-3} for better performance.

(\romannumeral1) For the network architecture, the important hyper-parameters include the dimensions of latent space $Z$, the dimensions of state features $d$, and the use of skip-connection between the prior and generative networks. We add an ablation study in Tab.~\ref{hyper-1} to perform a grid search. The result shows that (a) the latent space with relatively low-dimensions (i.e., $Z\in \mathbb{R}^{128}$) is sufficient to achieve good performance; (b) $d=512$ performs the best, which has much lower dimensions compared to the original image and results in a lower computational cost in VDM; and (c) skip-connection between the prior and generative networks is essential, and we have explained the reason theoretically in \S\ref{VDM-s2}. (\romannumeral2) For the learning rate in VDM optimization, we use the same learning rate (i.e., $1e-4$) as the basic PPO optimizer. Because the VDM network and actor-critic network make up a calculation graph together, using the same learning rate makes it convenient for training in practice. (\romannumeral3) For the training speed $t_{\rm vdm}$ in optimization, Fig.~\ref{fig:speed} shows that our default setting that $t_{\rm vdm}=3$ performs the best. Meanwhile, we find the performance of VDM is insensitive to this parameter.
(\romannumeral4) For the number of samples $k$ in the intrinsic reward $r^i_k$, a larger $k$ gives a tighter upper bound of $r^i$ according to Theorem \ref{theorem1}. We conduct experiments to compare the different settings of $k$ in Breakout and Seaquest from sticky Atari games. As shown in Fig.~\ref{fig_compare_s}, the learning curve is unstable when $k=1$, which often occurs in the middle stage of training. To understand such instability with small $k$, note that when the agent discovers new scenarios in exploration, the posterior of the latent variable $\mathbf{z}$ changes significantly. In this case, the variance of the posterior is high. Hence, using a single hidden variable sampled from the posterior produces a high deviation in the estimation, which negatively impacts policy learning. Upon experiments, we find $k=10$ sufficient to obtain satisfactory results.

\begin{table}[t]
  \caption{Ablation study of the network architecture in VDM}
  \centering
  \setlength{\tabcolsep}{3pt}{
  \begin{tabular}{c|ccc|ccc|cc}
    \toprule
    & \multicolumn{3}{c|}{Dimensions of $Z$}  & \multicolumn{3}{c|}{Dimensions of $d$} & \multicolumn{2}{c}{Skip-connection} \\\hline
     Task    & 64 & 128 & 256 & 256 & 512 & 1024 & yes & no \\
    \hline
	Alien    & 575.1 & \textbf{608.2} & 603.2  & 593.1 & \textbf{608.2} & 603.1
	 & \textbf{608.2} & 472.5 \\
	Breakout & 369.2 & 373.5 & \textbf{382.1}  & 351.9 & \textbf{373.5} & 361.4 & \textbf{373.5} & 298.5 \\
	Seaquest & 791.3 & \textbf{794.1} & 769.2  & 750.4 & \textbf{794.1} & 789.3 & \textbf{794.1} & 636.2 \\
    \bottomrule
  \end{tabular}}\label{hyper-1}
\vspace{-1em}
\end{table}

\begin{figure}[t]
\centering
\subfigure[Breakout]{\includegraphics[width=1.7in]{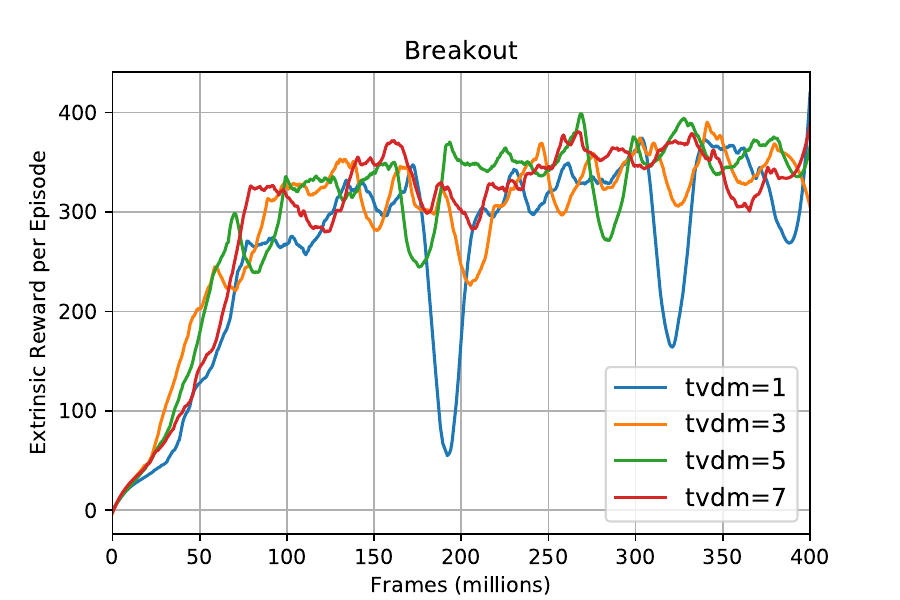}}
\subfigure[Seaquest]{\includegraphics[width=1.7in]{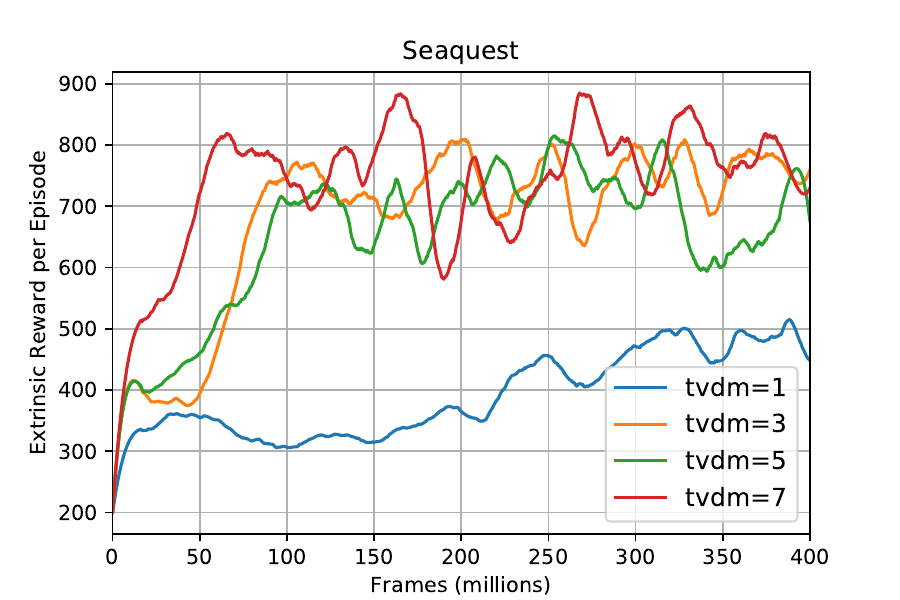}}
\caption{Ablation study of the \textbf{training speed} in VDM. $t_{\rm vdm}=1$ is not stable since the iteration of VDM is too slow for the changing policy, thus is hard to generate suitable intrinsic rewards. We find $t_{\rm vdm}=3, 5,$ and $7$ performs well. We choose to use $t_{\rm vdm}=3$ in experiments to obtain a good performance as well as the minimum computational cost.}
\label{fig:speed}
\vspace{-1em}
\end{figure}

\begin{figure}[t]
\centering
\subfigure[]{\includegraphics[width=1.6in]{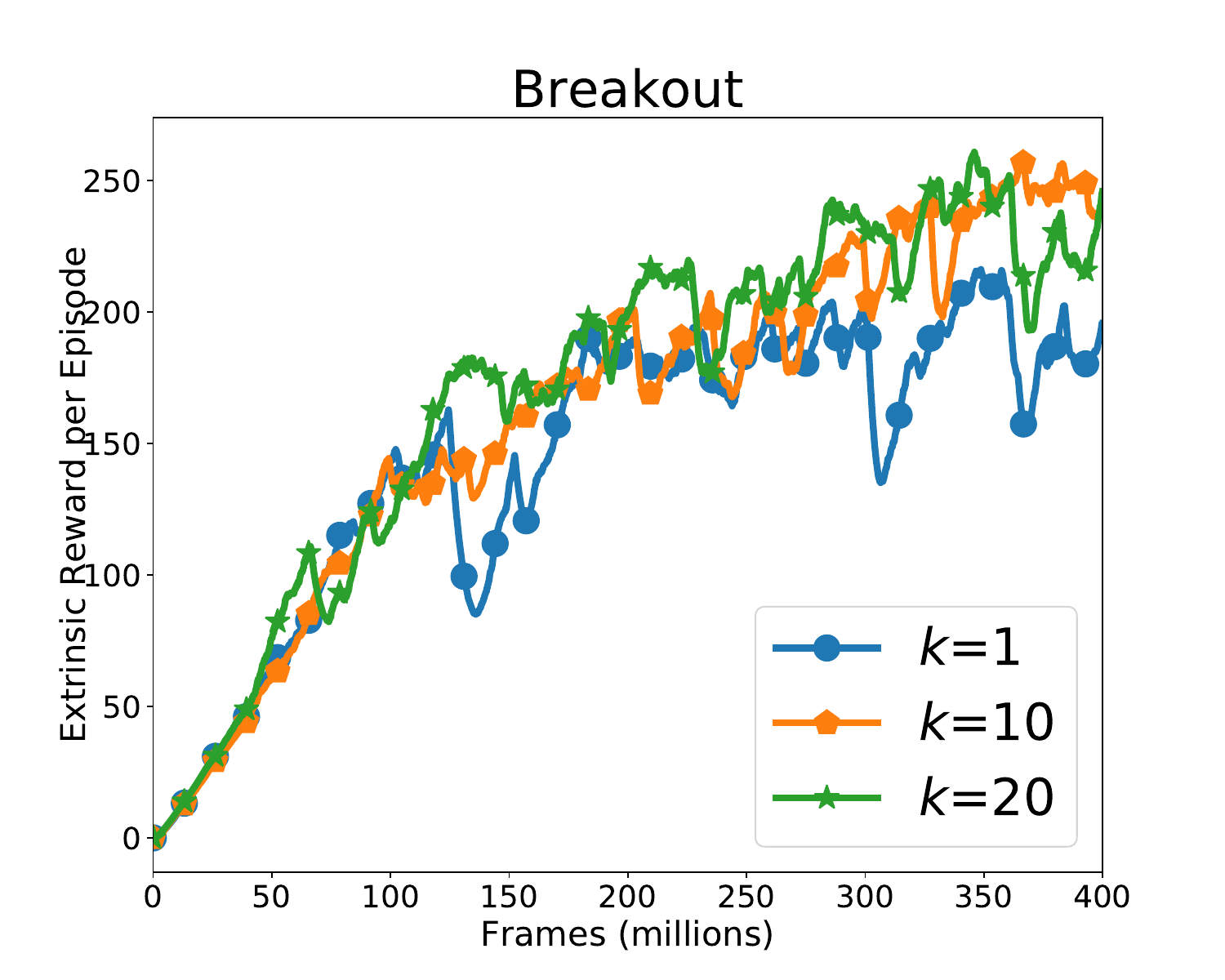}}
\subfigure[]{\includegraphics[width=1.6in]{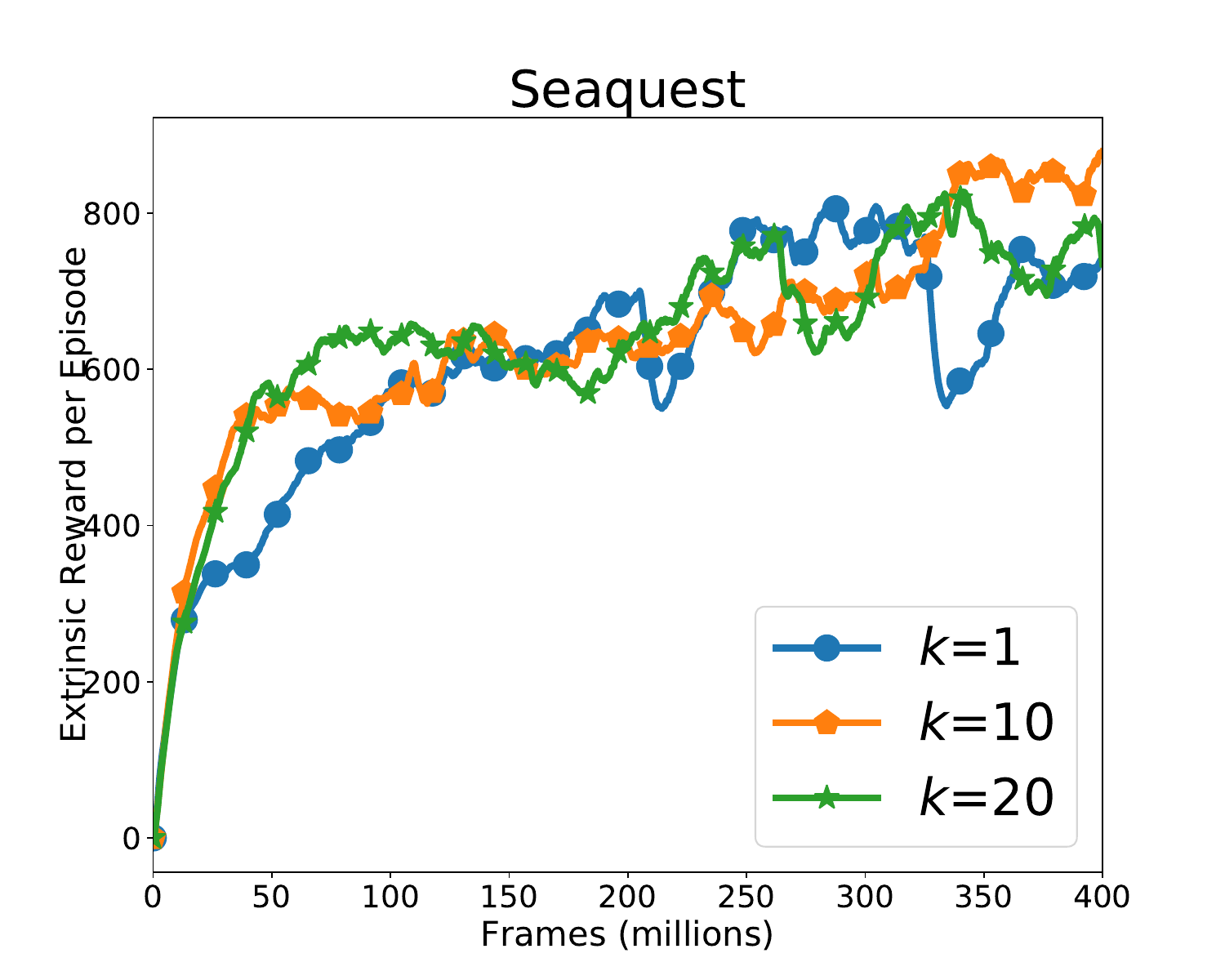}}
\caption{Result comparison of different settings of $k$ when calculating the intrinsic reward $r^i_k$ in sticky Atari games.}
\label{fig_compare_s}
\vspace{-0.5em}
\end{figure}

\section{Comparison with the basic variational model}

One may curious about whether other variational models can be used in exploration. In this section, we discuss the basic latent-variable models, i.e., variational auto-encoder (VAE), and applying Conditional VAE (CVAE) in modeling the multimodality and stochasticity of dynamics. Considering a typical VAE in modeling a random variable $X$ with the latent variable $z$, VAE has the variational objective as
\begin{equation}\nonumber
\mathcal{\tilde{L}}_{\rm VAE}=\mathbb{E}_{z\sim Q}[\log P(X|z)]-D_{\rm KL}[Q(z|X)\|P(Z)].
\end{equation}
where $P(Z)$ is set to be a standard Gaussian $N(0,1)$. CVAE modifies VAE by conditioning the generative process on an additional input $c$. The variational objective of CVAE is
\begin{equation}\nonumber
\mathcal{\tilde{L}}_{\rm CVAE}\!=\!\mathbb{E}_{z\sim Q(z|X,c)}[\log P(X|z,c)]\!-\!D_{\rm KL}[Q(z|X,c)\|N(0,1)],
\end{equation}
where $Q(z|X,c)$ servers as the encoder of CVAE, and $P(X|z,c)$ servers as the decoder.

\subsubsection{CVAE for Dynamics Modeling in RL} We aim to model the distribution of the next-state $s'$ conditioning on the current state-action pair $(s,a)$. Thus, $s'$ should be the target variable $X$, and $(s,a)$ should be the additional condition $c$ in CVAE. By substituting $X$ with $s'$, and substituting $c$ with $(s,a)$, we have the following objective in dynamics modeling as
\begin{equation}\label{eq:cvae}
\begin{aligned}
L_{\rm CVAE}=\mathbb{E}_{z\sim Q(z|s,a,s')}&[\log P(s'|s,a,z)]\\
&-D_{\rm KL}[Q(z|s,a,s')\|N(0,1)],
\end{aligned}
\end{equation}
where the encoder of CVAE-dynamics is $Q(z|s,a,s')$ and the decoder is $P(s'|s,a,z)$.

\begin{figure}[t]
  \centering
  \includegraphics[width=0.40\textwidth]{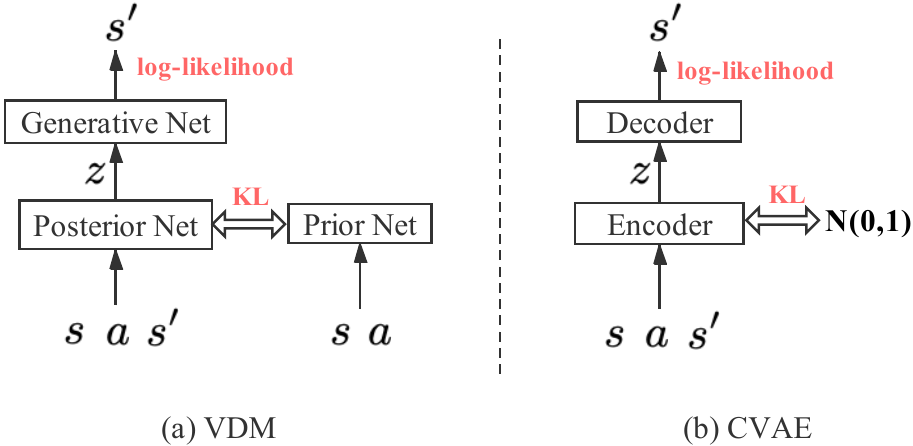}
  \caption{The architecture comparison between (a) the proposed VDM and (b) a basic variational methods (i.e., CVAE).}
  \label{fig:cvae-vdm}
  \vspace{-1em}
\end{figure}

\subsubsection{Theoretical comparison between CVAE and VDM} To illustrate the difference between CVAE and VDM, we simplify the model details and compare the architecture of CVAE and VDM in Fig.~\ref{fig:cvae-vdm}. We find the encoder in CVAE is similar to the posterior network in VDM, and the decoder in CVAE is similar to the generative network in VDM. The CVAE architecture does not include a prior network that encodes the information of $(s,a)$, and instead fixes the prior as a standard Gaussian. Comparing $L_{\rm VDM}$ in \eqref{eq-lb} and the objective of CVAE in \eqref{eq:cvae}, the differences are as follows.

(\romannumeral1) According to the KL-divergence term, CVAE aims to push the posterior of $Z$ to $N(0,1)$, while VDM aims to push $Z$ to a changing prior $p(z|s,a)$ that is trained along with the neural network. Usually, the KL-term in Eq.~\eqref{eq-lb} is less then KL-term in Eq.~\eqref{eq:cvae}, as
\begin{equation}\label{eq:ana-1}
D_{\rm KL}[q(z|s,a,s') \| p(z|s,a)]\leq D_{\rm KL}[Q(z|s,a,s') \| N(0,1)].
\end{equation}
The reason is that the prior $p(z|s,a)$ of VDM encodes the information of underlying MDP in training, while the prior of CVAE is fixed and does not contain any information. Thus the KL-divergence between the posterior and VDM-prior is easier to minimize than CVAE-prior.

(\romannumeral2) The sampling distributions of $\mathbb{E}_{z}[\log p(s'|s,a,z)]$ are different. The KL-divergence of CVAE is a stronger regularization term since the posterior needs to approach a fixed prior. As result, the log-likelihood of the next-state in CVAE is harder to maximize compared to VDM. Then we have
\begin{equation}\label{eq:ana-2}
\mathbb{E}_{q(z|s,a,s')}[\log p(s'|s,a,z)]\geq \mathbb{E}_{Q(z|s,a,s')}[\log P(s'|s,a,z)].
\end{equation}

Following \eqref{eq:ana-1}, \eqref{eq:ana-2}, and $\log p(s'|s,a)\geq L_{\rm VDM}$, we have
\begin{equation}
\log p(s'|s,a)\geq L_{\rm VDM} \geq L_{\rm CVAE}.
\end{equation}
As a result, the objective of VDM is a \emph{tighter} bound of dynamics' log-likelihood than CAVE. Theoretically, our method can better approximate the dynamics of the underlying MDP.

\subsubsection{Empirically comparison} We implement a CVAE-based exploration algorithm by modifying the prior of VDM to a standard Gaussian\footnote{The code is released at \url{https://github.com/Baichenjia/CAVE_NoisyMinist} (for Noisy-Mnist) and \url{https://github.com/Baichenjia/CVAE_exploration} (for other tasks) for reproducibility and further improvement.}. For Noisy-Mnist, we train the model for 200 epochs, and the result is shown in Fig.~\ref{fig:cvae-mnist}. From the result, we find the CVAE can predict a part of next-state classes like "1, 2, 3, 7, 8, 9" with various writing styles, however, the quality and diversity of prediction is worse than VDM. For Atari, we run experiments in two popular Atari games: Alien and Breakout. The result is shown in Fig.~\ref{fig:cvae-atari}. Surprisingly, we find CVAE performs as well as VDM in Alien. The possible reason is that the dynamics are relatively simple in this task, thus the strong regularization does not affect the learning of dynamics. However, in Breakout, we find CVAE performs significantly worse than VDM. The theoretical benefit of VDM makes it better approximates the complex dynamics of ball and bricks in this task. 

\begin{figure}[t]
  \centering
  \includegraphics[width=0.4\textwidth]{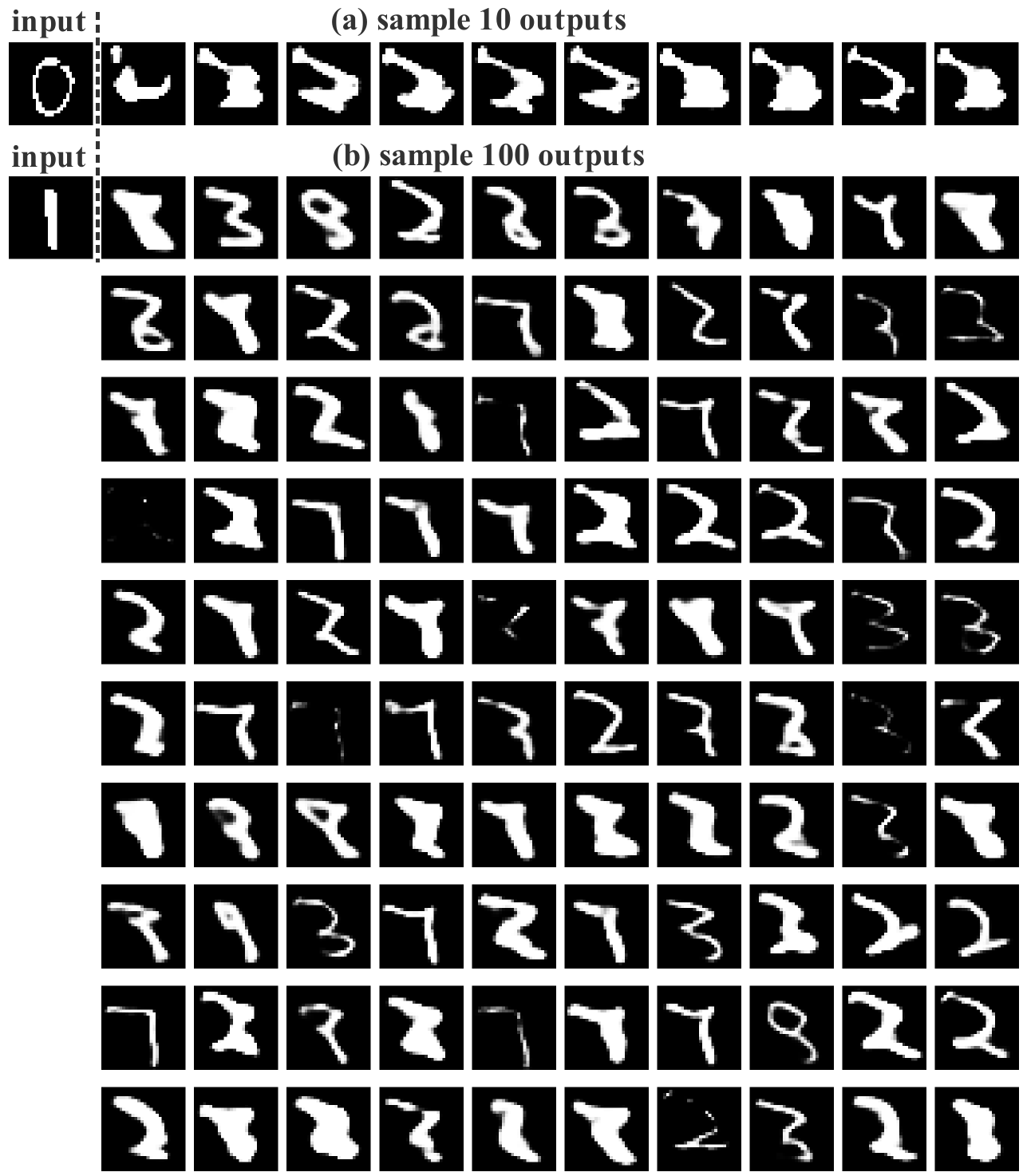}
\caption{Result of the CVAE-based dynamics model in `Noisy-Mnist'.}
\label{fig:cvae-mnist}
\vspace{-0.5em}
\end{figure}

To conclude, both in theory and in practice, VDM has advantages over the CVAE-based model.

\begin{figure}[t]
  \centering
  \includegraphics[width=0.43\textwidth]{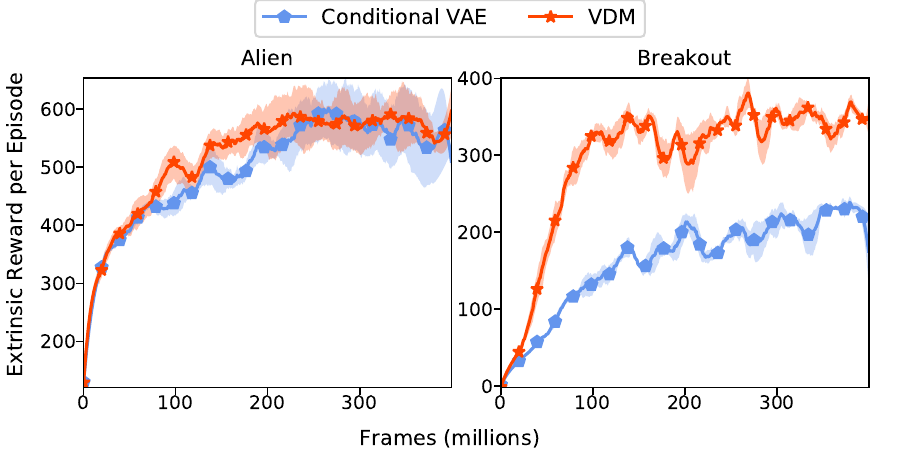}
\caption{Comparison of VDM-based and CVAE-based exploration in Atari.}
\label{fig:cvae-atari}
\vspace{-1em}
\end{figure}

\section{Conclusion}

In this paper, we use a variational dynamic model for self-supervised exploration in unknown environments. The self-exploratory agent does not require shaped or carefully hand-crafted rewards. We propose VDM to model the multimodality and stochasticity of the dynamic explicitly. To train VDM, we propose a variational learning objective. We further estimate an intrinsic reward based on VDM for exploration. Extensive experiments are conducted to compare VDM with various baseline methods. We observe that the exploratory policy performs well in dealing with the multimodality in Atari games and the additional stochasticity in sticky Atari games. The policies learned by VDM have good adaptability in the novel scenarios and achieve Nash equilibrium more efficiently. Moreover, we observe that VDM is promising in the real-world robotic task. We compare VDM with CVAE method both in theory and in practice, and find VDM has advantages over the CVAE-based model. Future works include combining VDM with extrinsic rewards and investigating VDM to off-policy algorithms.

% or


%\appendices

%\section{}

%\section*{Acknowledgment}
%
%The authors would like to thank Dr. Yiran Xue for helping with the UR5 robot-arm.

\ifCLASSOPTIONcaptionsoff
  \newpage
\fi

\bibliographystyle{IEEEtran}
\bibliography{IEEEabrv, main}

\begin{thebibliography}{10}
\providecommand{\url}[1]{#1}
\csname url@samestyle\endcsname
\providecommand{\newblock}{\relax}
\providecommand{\bibinfo}[2]{#2}
\providecommand{\BIBentrySTDinterwordspacing}{\spaceskip=0pt\relax}
\providecommand{\BIBentryALTinterwordstretchfactor}{4}
\providecommand{\BIBentryALTinterwordspacing}{\spaceskip=\fontdimen2\font plus
\BIBentryALTinterwordstretchfactor\fontdimen3\font minus
  \fontdimen4\font\relax}
\providecommand{\BIBforeignlanguage}[2]{{%
\expandafter\ifx\csname l@#1\endcsname\relax
\typeout{** WARNING: IEEEtran.bst: No hyphenation pattern has been}%
\typeout{** loaded for the language `#1'. Using the pattern for}%
\typeout{** the default language instead.}%
\else
\language=\csname l@#1\endcsname
\fi
#2}}
\providecommand{\BIBdecl}{\relax}
\BIBdecl

\bibitem{sutton-2018}
R.~S. Sutton and A.~G. Barto, \emph{Reinforcement learning: An
  introduction}.\hskip 1em plus 0.5em minus 0.4em\relax MIT press, 2018.

\bibitem{DQN-2015}
V.~Mnih, K.~Kavukcuoglu, D.~Silver, A.~A. Rusu, J.~Veness, M.~G. Bellemare,
  A.~Graves, M.~A. Riedmiller, A.~Fidjeland, G.~Ostrovski, S.~Petersen,
  C.~Beattie, A.~Sadik, I.~Antonoglou, H.~King, D.~Kumaran, D.~Wierstra,
  S.~Legg, and D.~Hassabis, ``Human-level control through deep reinforcement
  learning,'' \emph{Nature}, vol. 518, pp. 529--533, 2015.

\bibitem{ieee-2}
Z.~Ren, D.~Dong, H.~Li, and C.~Chen, ``Self-paced prioritized curriculum
  learning with coverage penalty in deep reinforcement learning,'' \emph{IEEE
  transactions on neural networks and learning systems}, vol.~29, no.~6, pp.
  2216--2226, 2018.

\bibitem{AlphaGo-2017}
D.~Silver, J.~Schrittwieser, K.~Simonyan, I.~Antonoglou, A.~Huang, A.~Guez,
  T.~Hubert, L.~R. Baker, M.~Lai, A.~Bolton, Y.~Chen, T.~P. Lillicrap, F.~Hui,
  L.~Sifre, G.~van~den Driessche, T.~Graepel, and D.~Hassabis, ``Mastering the
  game of go without human knowledge,'' \emph{Nature}, vol. 550, pp. 354--359,
  2017.

\bibitem{AlphaStar-2019}
O.~Vinyals, I.~Babuschkin, W.~M. Czarnecki, M.~Mathieu, A.~Dudzik, J.~Chung,
  D.~H. Choi, R.~Powell, T.~Ewalds, P.~Georgiev \emph{et~al.}, ``Grandmaster
  level in starcraft ii using multi-agent reinforcement learning,''
  \emph{Nature}, pp. 1--5, 2019.

\bibitem{liu2020generating}
P.~Liu, C.~Bai, Y.~Zhao, C.~Bai, W.~Zhao, and X.~Tang, ``Generating attentive
  goals for prioritized hindsight reinforcement learning,''
  \emph{Knowledge-Based Systems}, vol. 203, p. 106140, 2020.

\bibitem{ieee-3}
Z.~Yang, K.~Merrick, L.~Jin, and H.~A. Abbass, ``Hierarchical deep
  reinforcement learning for continuous action control,'' \emph{IEEE
  transactions on neural networks and learning systems}, vol.~29, no.~11, pp.
  5174--5184, 2018.

\bibitem{bai-her}
C.~Bai, L.~Wang, Y.~Wang, Z.~Wang, R.~Zhao, C.~Bai, and P.~Liu, ``Addressing
  hindsight bias in multigoal reinforcement learning,'' \emph{IEEE Transactions
  on Cybernetics}, pp. 1--14, 2021.

\bibitem{intrinsic-2000}
R.~M. Ryan and E.~L. Deci, ``Intrinsic and extrinsic motivations: Classic
  definitions and new directions,'' \emph{Contemporary educational psychology},
  vol.~25, no.~1, pp. 54--67, 2000.

\bibitem{curiosity-2017}
D.~Pathak, P.~Agrawal, A.~A. Efros, and T.~Darrell, ``Curiosity-driven
  exploration by self-supervised prediction,'' in \emph{Proceedings of
  International Conference on Machine Learning}, 2017, pp. 2778--2787.

\bibitem{largescale-2019}
Y.~Burda, H.~Edwards, D.~Pathak, A.~Storkey, T.~Darrell, and A.~A. Efros,
  ``Large-scale study of curiosity-driven learning,'' in \emph{Proceedings of
  International Conference on Learning Representations}, 2019, pp. 1--14.

\bibitem{trials-2018}
K.~Chua, R.~Calandra, R.~McAllister, and S.~Levine, ``Deep reinforcement
  learning in a handful of trials using probabilistic dynamics models,'' in
  \emph{Proceedings of Advances in Neural Information Processing Systems},
  2018, pp. 4754--4765.

\bibitem{disagree-2019}
D.~Pathak, D.~Gandhi, and A.~Gupta, ``Self-supervised exploration via
  disagreement,'' in \emph{Proceedings of International Conference on Machine
  Learning}, 2019, pp. 5062--5071.

\bibitem{vae-2014}
D.~P. Kingma and M.~Welling, ``Auto-encoding variational bayes,'' in
  \emph{Proceedings of International Conference on Learning Representations},
  2014, pp. 1--14.

\bibitem{cvae-2015}
K.~Sohn, H.~Lee, and X.~Yan, ``Learning structured output representation using
  deep conditional generative models,'' in \emph{Proceedings of Advances in
  neural information processing systems}, 2015, pp. 3483--3491.

\bibitem{sutton-2000}
R.~S. Sutton, D.~A. McAllester, S.~P. Singh, and Y.~Mansour, ``Policy gradient
  methods for reinforcement learning with function approximation,'' in
  \emph{Proceedings of Advances in neural information processing systems},
  2000, pp. 1057--1063.

\bibitem{trpo-2015}
J.~Schulman, S.~Levine, P.~Abbeel, M.~Jordan, and P.~Moritz, ``Trust region
  policy optimization,'' in \emph{Proceedings of International conference on
  machine learning}, 2015, pp. 1889--1897.

\bibitem{ppo-2017}
J.~Schulman, F.~Wolski, P.~Dhariwal, A.~Radford, and O.~Klimov, ``Proximal
  policy optimization algorithms,'' \emph{arXiv preprint arXiv:1707.06347},
  2017.

\bibitem{gae-2016}
J.~Schulman, P.~Moritz, S.~Levine, M.~Jordan, and P.~Abbeel, ``High-dimensional
  continuous control using generalized advantage estimation,'' in
  \emph{Proceedings of International Conference on Learning Representations},
  2016, pp. 1--14.

\bibitem{count-2017}
G.~Ostrovski, M.~G. Bellemare, A.~van~den Oord, and R.~Munos, ``Count-based
  exploration with neural density models,'' in \emph{Proceedings of
  International Conference on Machine Learning}, 2017, pp. 2721--2730.

\bibitem{count-2016}
M.~Bellemare, S.~Srinivasan, G.~Ostrovski, T.~Schaul, D.~Saxton, and R.~Munos,
  ``Unifying count-based exploration and intrinsic motivation,'' in
  \emph{Proceedings of Advances in Neural Information Processing Systems},
  2016, pp. 1471--1479.

\bibitem{reach-2019}
N.~Savinov, A.~Raichuk, R.~Marinier, D.~Vincent, M.~Pollefeys, T.~Lillicrap,
  and S.~Gelly, ``Episodic curiosity through reachability,'' in
  \emph{Proceedings of International Conference on Learning Representations},
  2019, pp. 1--14.

\bibitem{RND-2019}
Y.~Burda, H.~Edwards, A.~Storkey, and O.~Klimov, ``Exploration by random
  network distillation,'' in \emph{Proceedings of International Conference on
  Learning Representations}, 2019, pp. 1--14.

\bibitem{ngu-2020}
A.~P. Badia, P.~Sprechmann, A.~Vitvitskyi, D.~Guo, B.~Piot, S.~Kapturowski,
  O.~Tieleman, M.~Arjovsky, A.~Pritzel, A.~Bolt, and C.~Blundell, ``Never give
  up: Learning directed exploration strategies,'' in \emph{Proceedings of
  International Conference on Learning Representations}, 2020, pp. 1--14.

\bibitem{aware-2018}
N.~Haber, D.~Mrowca, S.~Wang, L.~F. Fei-Fei, and D.~L. Yamins, ``Learning to
  play with intrinsically-motivated, self-aware agents,'' in \emph{Proceedings
  of Advances in Neural Information Processing Systems}, 2018, pp. 8388--8399.

\bibitem{MAX-2019}
P.~Shyam, W.~Ja{\'s}kowski, and F.~Gomez, ``Model-based active exploration,''
  in \emph{Proceedings of International Conference on Machine Learning}, 2019,
  pp. 5779--5788.

\bibitem{bai-1}
C.~Bai, L.~Wang, L.~Han, J.~Hao, A.~Garg, P.~Liu, and Z.~Wang, ``Principled
  exploration via optimistic bootstrapping and backward induction,'' in
  \emph{Proceedings of International Conference on Machine Learning}, vol.
  139.\hskip 1em plus 0.5em minus 0.4em\relax PMLR, 2021, pp. 577--587.

\bibitem{sid-2019}
J.~Zhang, N.~Wetzel, N.~Dorka, J.~Boedecker, and W.~Burgard, ``Scheduled
  intrinsic drive: A hierarchical take on intrinsically motivated
  exploration,'' \emph{arXiv preprint arXiv:1903.07400}, 2019.

\bibitem{mulex-2019}
L.~Beyer, D.~Vincent, O.~Teboul, S.~Gelly, M.~Geist, and O.~Pietquin, ``Mulex:
  Disentangling exploitation from exploration in deep rl,'' in
  \emph{Proceedings of Exploration in RL Workshop of International Conference
  on Machine Learning}, 2019, pp. 1--14.

\bibitem{emi-2019}
H.~Kim, J.~Kim, Y.~Jeong, S.~Levine, and H.~O. Song, ``Emi: Exploration with
  mutual information,'' in \emph{Proceedings of International Conference on
  Machine Learning}, 2019, pp. 3360--3369.

\bibitem{kim2019curiosity}
Y.~Kim, W.~Nam, H.~Kim, J.-H. Kim, and G.~Kim, ``Curiosity-bottleneck:
  Exploration by distilling task-specific novelty,'' in \emph{Proceedings of
  International Conference on Machine Learning}.\hskip 1em plus 0.5em minus
  0.4em\relax PMLR, 2019, pp. 3379--3388.

\bibitem{DB-2021}
C.~Bai, L.~Wang, L.~Han, A.~Garg, J.~Hao, P.~Liu, and Z.~Wang, ``Dynamic
  bottleneck for robust self-supervised exploration,'' in \emph{Proceedings of
  Advances in Neural Information Processing Systems}, 2021, pp. 1--14.

\bibitem{choi2018contingency}
J.~Choi, Y.~Guo, M.~Moczulski, J.~Oh, N.~Wu, M.~Norouzi, and H.~Lee,
  ``Contingency-aware exploration in reinforcement learning,'' in
  \emph{Proceedings of International Conference of Learning Representation},
  2019, pp. 1--14.

\bibitem{vi-1}
D.~M. Blei, A.~Kucukelbir, and J.~D. McAuliffe, ``Variational inference: A
  review for statisticians,'' \emph{Journal of the American statistical
  Association}, vol. 112, no. 518, pp. 859--877, 2017.

\bibitem{vi-2}
M.~Toussaint and A.~Storkey, ``Probabilistic inference for solving discrete and
  continuous state markov decision processes,'' in \emph{Proceedings of
  International Conference on Machine Learning}, 2006, pp. 945--952.

\bibitem{vi-3}
B.~D. Ziebart, A.~L. Maas, J.~A. Bagnell, and A.~K. Dey, ``Maximum entropy
  inverse reinforcement learning.'' in \emph{Proceedings of the AAAI Conference
  on Artificial Intelligence}, vol.~8.\hskip 1em plus 0.5em minus 0.4em\relax
  Chicago, IL, USA, 2008, pp. 1433--1438.

\bibitem{chen2019variational}
L.~Chen, L.~Wang, Z.~Han, J.~Zhao, and W.~Wang, ``Variational inference based
  kernel dynamic bayesian networks for construction of prediction intervals for
  industrial time series with incomplete input,'' \emph{IEEE/CAA Journal of
  Automatica Sinica}, vol.~7, no.~5, pp. 1437--1445, 2019.

\bibitem{vi-4}
P.~Dayan and G.~E. Hinton, ``Using expectation-maximization for reinforcement
  learning,'' \emph{Neural Computation}, vol.~9, no.~2, pp. 271--278, 1997.

\bibitem{vi-5}
J.~Peters and S.~Schaal, ``Reinforcement learning by reward-weighted regression
  for operational space control,'' in \emph{Proceedings of International
  Conference on Machine Learning}, 2007, pp. 745--750.

\bibitem{vi-6}
S.~Levine, \emph{Motor skill learning with local trajectory methods}.\hskip 1em
  plus 0.5em minus 0.4em\relax Stanford University, 2014.

\bibitem{vi-7}
M.~Fellows, A.~Mahajan, T.~G. Rudner, and S.~Whiteson, ``Virel: A variational
  inference framework for reinforcement learning,'' \emph{Advances in Neural
  Information Processing Systems}, vol.~32, pp. 7122--7136, 2019.

\bibitem{vi-8}
D.~P. Bertsekas, ``Feature-based aggregation and deep reinforcement learning: A
  survey and some new implementations,'' \emph{IEEE/CAA Journal of Automatica
  Sinica}, vol.~6, no.~1, pp. 1--31, 2018.

\bibitem{vi-9}
X.~Sun and B.~Bischl, ``Tutorial and survey on probabilistic graphical model
  and variational inference in deep reinforcement learning,'' in
  \emph{Proceedings of 2019 IEEE Symposium Series on Computational Intelligence
  (SSCI)}.\hskip 1em plus 0.5em minus 0.4em\relax IEEE, 2019, pp. 110--119.

\bibitem{vime-2016}
R.~Houthooft, X.~Chen, Y.~Duan, J.~Schulman, F.~De~Turck, and P.~Abbeel,
  ``Vime: Variational information maximizing exploration,'' in
  \emph{Proceedings of Advances in Neural Information Processing Systems},
  2016, pp. 1109--1117.

\bibitem{vi-10}
D.~Corneil, W.~Gerstner, and J.~Brea, ``Efficient model-based deep
  reinforcement learning with variational state tabulation,'' in
  \emph{Proceedings of International Conference on Machine Learning}.\hskip 1em
  plus 0.5em minus 0.4em\relax PMLR, 2018, pp. 1049--1058.

\bibitem{schulz2018stochastic}
P.~Schulz, W.~Aziz, and T.~Cohn, ``A stochastic decoder for neural machine
  translation,'' in \emph{The 56th Annual Meeting of the Association for
  Computational Linguistics (Volume 1: Long Papers)}, 2018, pp. 1243--1252.

\bibitem{vaeac-2019}
O.~Ivanov, M.~Figurnov, and D.~Vetrov, ``Variational autoencoder with arbitrary
  conditioning,'' in \emph{Proceedings of International Conference on Learning
  Representations}, 2019, pp. 1--14.

\bibitem{chua2018deep}
K.~Chua, R.~Calandra, R.~McAllister, and S.~Levine, ``Deep reinforcement
  learning in a handful of trials using probabilistic dynamics models,'' in
  \emph{Proceedings of Advances in Neural Information Processing Systems},
  2018, pp. 4759--4770.

\bibitem{jalr-2018}
M.~C. Machado, M.~G. Bellemare, E.~Talvitie, J.~Veness, M.~Hausknecht, and
  M.~Bowling, ``Revisiting the arcade learning environment: Evaluation
  protocols and open problems for general agents,'' \emph{Journal of Artificial
  Intelligence Research}, vol.~61, pp. 523--562, 2018.

\bibitem{sekar2020planning}
R.~Sekar, O.~Rybkin, K.~Daniilidis, P.~Abbeel, D.~Hafner, and D.~Pathak,
  ``Planning to explore via self-supervised world models,'' in
  \emph{Proceedings of International Conference on Machine Learning}, 2020, pp.
  8583--8592.

\bibitem{stable-2}
T.~Karras, T.~Aila, S.~Laine, and J.~Lehtinen, ``Progressive growing of gans
  for improved quality, stability, and variation,'' in \emph{Proceedings of
  International Conference of Learning Representation}, 2018, pp. 1--14.

\bibitem{stable-3}
H.~Thanh-Tung, T.~Tran, and S.~Venkatesh, ``Improving generalization and
  stability of generative adversarial networks,'' in \emph{Proceedings of
  International Conference of Learning Representation}, 2019, pp. 1--14.

\bibitem{exp-1}
S.~Gao, M.~Zhou, Y.~Wang, J.~Cheng, H.~Yachi, and J.~Wang, ``Dendritic neuron
  model with effective learning algorithms for classification, approximation,
  and prediction,'' \emph{IEEE transactions on neural networks and learning
  systems}, vol.~30, no.~2, pp. 601--614, 2018.

\bibitem{exp-2}
J.~Wang and T.~Kumbasar, ``Parameter optimization of interval type-2 fuzzy
  neural networks based on pso and bbbc methods,'' \emph{IEEE/CAA Journal of
  Automatica Sinica}, vol.~6, no.~1, pp. 247--257, 2019.

\bibitem{exp-3}
P.~Zhang, S.~Shu, and M.~Zhou, ``An online fault detection method based on
  svm-grid for cloud computing systems,'' \emph{IEEE/CAA J. Automat. Sinica},
  vol.~5, no.~2, pp. 445--456, 2018.

\end{thebibliography}

\end{document}